\documentclass[gmd, manuscript]{copernicus}

\usepackage[final]{pdfpages}
\usepackage{amsfonts}
\usepackage{amsmath}
\usepackage{color}
\usepackage{url}
\usepackage{wrapfig}
\usepackage{float}
\usepackage{mdwlist}
\usepackage{subfig}
\usepackage{placeins}
\floatstyle{boxed} 
\usepackage{pdfpages}
\usepackage{lineno}
\usepackage{comment}
\usepackage{hyperref}
\usepackage{lineno}
\usepackage[compact]{titlesec}
\usepackage{outlines}

\usepackage{booktabs}
\usepackage[table]{xcolor}
\newcommand{\midsepremove}{\aboverulesep = 0mm \belowrulesep = 0mm}
\usepackage{pifont}
\newcommand{\cmark}{\ding{51}}%
\newcommand{\xmark}{\ding{55}}%

\nolinenumbers

\begin{document}

\title{Huge Ensembles Part II: Properties of a Huge Ensemble of Hindcasts Generated with Spherical Fourier Neural Operators}


\Author[1,2,*][ankur.mahesh@berkeley.edu]{Ankur}{Mahesh} 
\Author[1,2,*]{William}{D. Collins}
\Author[3]{Boris}{Bonev}
\Author[3]{Noah}{Brenowitz}
\Author[3]{Yair}{Cohen}
\Author[4]{Peter}{Harrington}
\Author[3]{Karthik}{Kashinath}
\Author[3]{Thorsten}{Kurth}
\Author[1]{Joshua}{North}
\Author[5,1]{Travis A.}{O'Brien}
\Author[3,6]{Michael}{Pritchard}
\Author[3]{David}{Pruitt}
\Author[1]{Mark}{Risser}
\Author[4]{Shashank}{Subramanian}
\Author[4]{Jared}{Willard}
\affil[1]{Earth and Environmental Sciences Area, Lawrence Berkeley National Laboratory (LBNL), Berkeley, California, USA}
\affil[2]{Department of Earth and Planetary Science, University of California, Berkeley, USA}
\affil[3]{NVIDIA Corporation, Santa Clara, California, USA}
\affil[4]{National Energy Research Scientific Computing Center (NERSC), LBNL, Berkeley, California, USA}
\affil[5]{Department of Earth and Atmospheric Sciences, Indiana University, Bloomington, Indiana, USA}
\affil[6]{Department of Earth System Science, University of California, Irvine, USA}
\affil[*]{These authors contributed equally to this work.}

\runningtitle{GENERATING HUGE ENSEMBLES USING SFNO: Part II}

\runningauthor{MAHESH ET AL}

\received{TBD}
\pubdiscuss{} 
\revised{TBD}
\accepted{TBD}
\published{TBD}


\firstpage{1}

\maketitle

\begin{abstract}
In Part~I, we created an ensemble based on Spherical Fourier Neural Operators. As initial condition perturbations, we used bred vectors, and as model perturbations, we used multiple checkpoints trained independently from scratch.  Based on diagnostics that assess the ensemble's physical fidelity, our ensemble has comparable performance to operational weather forecasting systems.  However, it requires orders of magnitude fewer computational resources.  Here in Part II, we generate a huge ensemble (HENS), with 7,424 members initialized each day of summer 2023.  We enumerate the technical requirements for running huge ensembles at this scale. HENS precisely samples the tails of the forecast distribution and presents a detailed sampling of internal variability. HENS has two primary applications: (1) as a large dataset with which to study the statistics and drivers of extreme weather and (2) as a weather forecasting system. For extreme climate statistics, HENS samples events 4$\sigma$ away from the ensemble mean. At each grid cell, HENS increases the skill of the most accurate ensemble member and enhances coverage of possible future trajectories.  As a weather forecasting model, HENS issues extreme weather forecasts with better uncertainty quantification. It also reduces the probability of outlier events, in which the verification value lies outside the ensemble forecast distribution.


\end{abstract}

\section{Introduction}

Ensemble forecasts are an invaluable tool in weather and climate forecasting. By characterizing the probability distribution of possible future outcomes, they improve decision-making in the face of uncertainty \citep{Mankin2020}.  In operational numerical weather prediction, there has been significant progress in accurately representing uncertainty to create probabilistic ensemble predictions \citep{Palmer2002, Leutbecher2008}.  In climate science, large ensembles are essential to separate internal variability from forced trends \citep{Deser2020, Kay2015}. Notable examples include the 100-member Community Earth System Model Large Ensemble and the 1000-member Observational Large Ensemble, which uses bootstrap resampling and signal processing methods to represent internal variability \citep{McKinnon2017}. 

Small sample sizes are a major challenge to characterizing and researching extremes in the observational record \citep{Thompson2017, Zhang2024, BercosHickey2022, Philip2022}. Low-likelihood high-impact events, such as heatwaves beyond 3 standard deviations away from the climatological mean (e.g. those characterized in \citet{Zhang2023}), have significant implications for human society and health. The observational record is limited to approximately fifty years, based on the start of the satellite era.  Ensembles of weather and climate simulations alleviate this challenge by providing a large sample size of plausible atmospheric states and trajectories.  \citet{Finkel2023} characterize sudden stratospheric warming events using a large dataset of subseasonal-to-seasonal ensemble hindcasts.   The UNprecedented Simulated Extreme ENsemble (UNSEEN) approach tests whether an ensemble prediction system is fit for purpose by assessing its stability and fidelity \citep{Kelder2022, Kelder2022IOP}.  After they have been validated, ensembles have been used to quantify risk \citep{Thompson2017}, return-time \citep{Leach2024}, trends \citep{KirchmeierYoung2020}, attribution to climate change \citep{Leach2021}, and future changes \citep{Swain2020} of low-likelihood, high-impact extremes.

Given the benefits of ensemble predictions, a core design decision is the number of members used in the ensemble.  The ensemble size has significant implications for the forecast's accuracy, uncertainty, and reliability. For climate simulations, \citet{Milinski2020} outline key steps and requirements to create initial condition ensembles.  Based on a user-specified threshold for acceptable error and uncertainty, they present a framework to calculate the required number of members.  In weather forecasting, \citet{Leutbecher2018} discuss the sampling uncertainty associated with finite sample sizes.  They also assess how probabilistic scores, such as the continuous ranked probability score (CRPS), converge as a function of ensemble size. Using reliability diagrams, \citet{Richardson2001} demonstrate that 50-member ensembles are more reliable than 20-member ones. \citet{Siegert2019} assess the effect of ensemble size if the forecast is converted to a normal distribution, using the mean and standard deviation from the ensemble.  In ensembles of up to 32 members, \citet{Buizza1998} characterize the effect of ensemble size on two metrics: (1) the skill of the best ensemble member at each grid point and (2) the outlier statistic. These two metrics reveal the benefit of larger ensembles, and we calculate these two statistics in a modern-day huge ensemble.

\begin{figure}[t]
\includegraphics[width=0.9\textwidth]{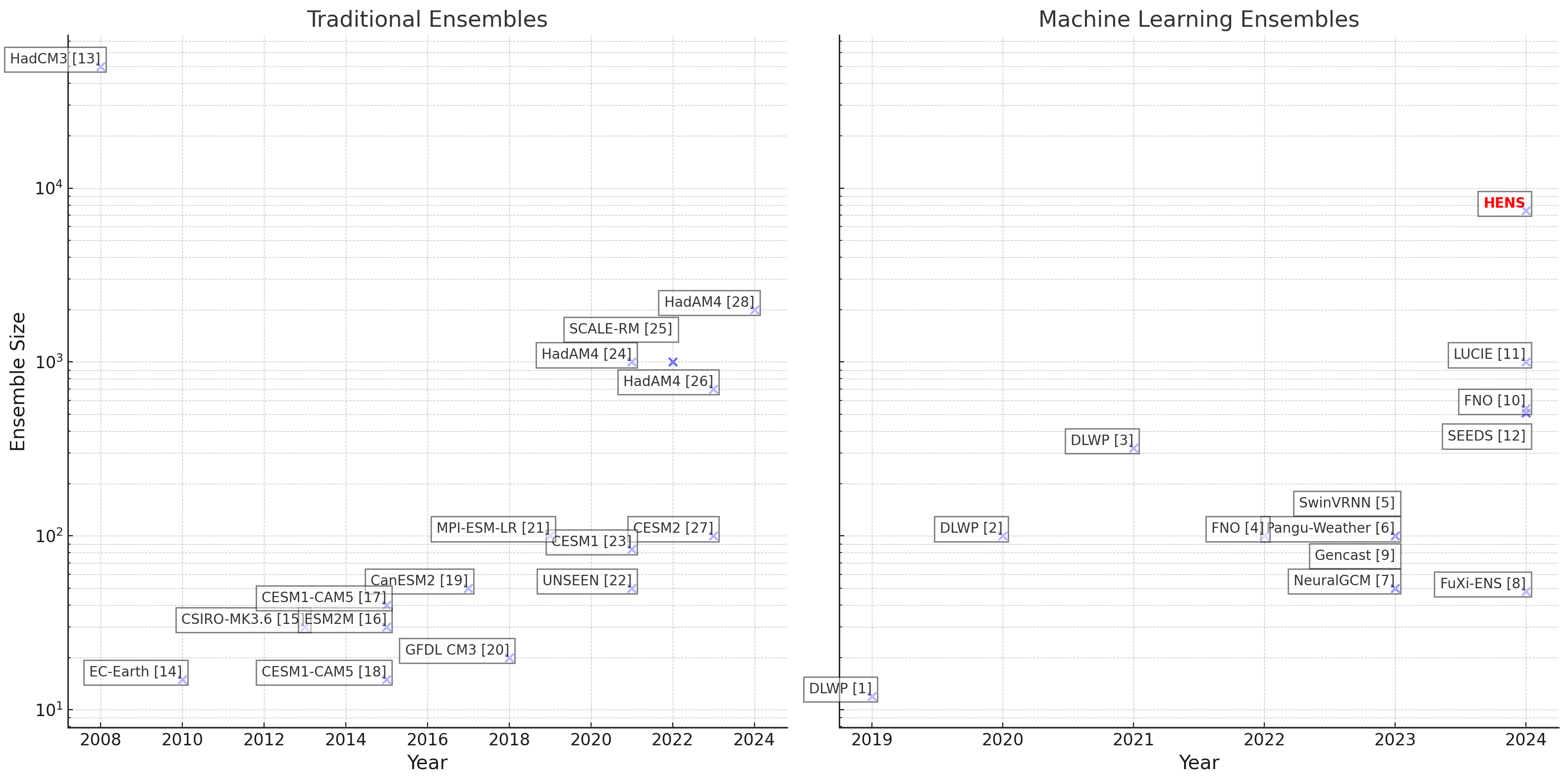}
\caption{\textbf{Ensemble Sizes in Weather and Climate Prediction}.   The left panel shows the ensemble size of traditional ensembles, which rely on numerical, physics-based simulation.  The right panel shows ensemble sizes from machine learning weather prediction ensembles.  Huge Ensembles (HENS) is the ensemble presented here, and it is highlighted with red. Bracketed numbers correspond to the numbered list of references for this figure provided in Section~\ref{app:ensemble_size_reference}.}
\label{fig:ensemble_size_time}
\end{figure}

Due to computational costs, it is impractical to run massive ensembles with numerical weather and climate models. In Part I \citep{MaheshPartI2024}, we introduced an ensemble based on Spherical Fourier Neural Operators, Bred Vectors, and Multiple Checkpoints trained from scratch (SFNO-BVMC).  The ensemble is orders of magnitude faster than comparable numerical simulations, so it enables the generation of huge ensembles (HENS).  We validated the fidelity of SFNO-BVMC extremes using mean, spectral, and extreme diagnostics. Here in Part II, we use SFNO-BVMC to generate ensemble hindcasts initialized on each day of summer 2023.  We choose summer 2023 as our test period because it is the hottest summer in the observed record \citep{esper20242023}.  Therefore, it is an important period to validate forecasts of extreme heatwaves and to analyze low-likelihood high-impact heatwaves in a warming world. We use SFNO-BVMC to generate a huge ensemble with 7,424 members; each ensemble member is run for 15 days.  Throughout this manuscript, we refer to this particular set of huge ensemble hindcasts during summer 2023 as "HENS."  We contextualize the size of HENS compared to other large ensembles in Figure~\ref{fig:ensemble_size_time}.  HENS~has more members than the majority of ensembles used in the past.

The central motivation for HENS is to remove the limitation of small sample size when studying and forecasting low-likelihood, high-impact extremes.  In~this manuscript, we present the HENS dataset and framework as a tool to generate many plausible (yet counterfactual) reconstructions of weather. This tool opens a variety of scientific questions, so we assess whether HENS provides a trustworthy sample of extreme weather events at the tails of the forecast distribution.  Since HENS offers a rich sampling of internal variability, we consider its utility for extreme statistics, and since HENS is based on an ML-based weather prediction model, we consider its utility for weather forecasting.

We present specific properties of HENS as a tool to study extremes.  To demonstrate the behavior across all of summer 2023, we calculate aggregate metrics, and we provide a case study for extreme climate statistics and for weather forecasting. In~this manuscript, we present three contributions:

\begin{enumerate}
    \item We list the technical requirements and considerations of creating ensemble sizes at this scale.
    \item As a large hindcast of simulated weather extremes, HENS enables improved climatic understanding of extreme weather events and their statistics.
    \begin{enumerate}
        \item HENS samples low-likelihood events at the tail of the forecast distribution.
        \item HENS provides a large sample size of counterfactual heat extremes.
        \item HENS produces better analogs to observed events by reducing the error of the best ensemble member
    \end{enumerate}
    \item We assess HENS's ability to forecast extreme heat events. 
    \begin{enumerate}
        \item HENS provides a finer sampling of the conditional forecast distribution, given that the forecast values exceed a climatological extreme threshold
        \item HENS reduces sampling uncertainty and has narrower confidence intervals for extreme temperature forecasts
        \item HENS captures heat extremes missed by smaller ensembles
    \end{enumerate}
\end{enumerate}

\section{Generating the Huge Ensemble}

\subsection{Technical Setup}

In Part I, we used bred vectors to represent initial condition uncertainty, and we trained twenty-nine SFNO checkpoints from scratch to represent model uncertainty. Here, we combine these two perturbation techniques to create a 7,424-member ensemble: 256 initial condition perturbations for each of the 29 SFNO checkpoints. Bred vectors, which are flow-dependent perturbations, are generated separately for each SFNO checkpoint and each forecast initial time.  


The HENS forecasts are initialized on each day of June, July, August 2023 at 00:00 UTC, for a total of 92 initial dates.  Each ensemble is run forward for 60 integration steps, or 360 hours.  In summary, our HENS simulation has the following dimensions:

\begin{enumerate}
    \item \textbf{Number of Saved Variables: } 12
    \item \textbf{Initial Time: } 92 initial days
    \item \textbf{Lead Time: } 61 time steps (One perturbed initial condition followed by a 360-hour forecast rollout)
    \item \textbf{Ensemble: } 7424 ensemble members
    \item \textbf{Latitude: } 721 (0.25 degrees)
    \item \textbf{Longitude: } 1440 (0.25 degrees)
\end{enumerate}

Generating and analyzing ensemble simulations at this scale is an important technical frontier. The latitude and longitude dimensions are based on the 0.25-degree horizontal resolution of European Center for Medium-range Weather Forecasts Reanalysis v5 (ERA5) \citep{Hersbach2020}, which is the SFNO-BVMC training dataset.  The HENS simulation output is saved in 32-bit floating-point format.  We store 12 variables and create a total dataset size of approximately 2 petabytes. Each variable in HENS thus takes 173 TB of space. Generating the ensemble for each initial date requires 21 TB of space and 45 minutes using 256 80GB NVIDIA A100 GPUs.  With 92 initial times, generating HENS costs 18,432 GPU-hours (256 GPUs for 3 days).  This quantity of data challenges the capabilities of existing climate model analysis workflows.  In Appendix~\ref{app:postprocessing}, we outline our post-processing strategy to analyze such large data volumes.   If we stored all the channels in the ensemble (not our chosen subset of 12 channels), then the total size of the dataset would be approximately 25 petabytes.

\begin{table}[]
\caption{\textbf{Subset of channels saved by the huge ensemble.} SFNO has 74 total prognostic channels (described in Part I). We save the following subset of channels in our HENS run.  Some of the saved channels are derived from on a combination of prognostic channels; they are calculated inline on the GPU during the ensemble generation. }
\rowcolors{2}{white}{gray!15}
\midsepremove
\begin{tabular}{llcc}
\toprule
\textbf{Type} & \textbf{Variable} & \textbf{Pressure Levels (hPa)} & \textbf{Prognostic} \\
\midrule
\cellcolor{white} & temperature & 850, 500 & \cmark\\
\cellcolor{white} \multirow{-2}{*}{Atmospheric Variables} & geopotential & 500, 300 & \cmark \\
\midrule
\cellcolor{white} & 2m air temperature & surface & \cmark\\
\cellcolor{white} & 2m dewpoint temperature & surface & \cmark\\
\cellcolor{white} & total column water vapor & surface & \cmark\\
\cellcolor{white} & sea level pressure & surface & \cmark\\
\cellcolor{white} \multirow{-5}{*}{Surface Variables} & surface pressure & surface  & \cmark  \\
\midrule
\cellcolor{white} & integrated vapor transport & - & \xmark\\
\cellcolor{white} & 2m heat index & surface & \xmark\\
\cellcolor{white} \multirow{-3}{*}{Derived} & 10m wind speed & surface & \xmark \\
\bottomrule
\end{tabular}
\label{table:variables}
\end{table}

To create the HENS hindcasts, we run SFNO in inference mode, in which Pytorch's automatic differentiation is turned off.  We use NVIDIA's earth2mip library for inference.  Our computations run on Perlmutter \citep{Perlmutter}, the high-performance computer at the National Energy Research Scientific Computing Center (NERSC).   For optimal I/O, the ensemble is written to scratch on Perlmutter's Lustre solid state scratch file system.  At the time of generation, our scratch disk storage allocation was 100~TB, which is not large enough to hold all 2 petabytes of the ensemble at once.  Scratch is meant to serve as a temporary high-bandwidth file storage solution, where data can be stored for up to 8 weeks.  Therefore, the HENS simulations need to be transferred to larger, long-term storage on NERSC's Community File System (CFS).   We use Globus to transfer the ensemble from scratch to CFS. Since Globus leverages load-optimized, parallel file transfers via GridFTP \citep{Ananthakrishnan2014}, it is much faster than native Linux commands for copying data. The transfers run on dedicated nodes on Perlmutter and occur concurrently with ensemble generation.   Due to the computational efficiency of ML weather forecasts, it is feasible to generate 256 ensemble members (each running on 1 GPU) in parallel per minute.  This introduces new data transfer considerations to ensure the data can be moved to its storage location in time.

We present an example of the ensemble generation workflow. We first generate the ensemble initialized on June 1, 2023 at 00:00 UTC.  After it completes, the Globus transfer from scratch to CFS begins.  While this transfer is occurring, we begin generating the huge ensemble initialized on June 2, 2023 at 00:00 UTC.  After the June 1 ensemble transfer completes, it is deleted from scratch. Crucially, the Globus transfers have 25 GB/s speeds, so they transfer each initial date's 21 TB ensemble in approximately 15 minutes.  With a native Linux command, the transfer took approximately 4 hours.  Unlike the Linux commands, Globus transfers the ensemble in less time than it takes to generate the ensemble (45 minutes).  Using simultaneous data generation and data transfer, we do not exceed our scratch disk storage allocation.  At any given time, no more than two initial days' worth of ensembles are stored on scratch. Our computational allocation goes entirely towards data generation rather than data transfer, and GPUs are not idle while the transfer from scratch to CFS occurs.

\subsection{Regenerating the Ensemble}

Traditionally, climate simulations and weather hindcasts are stored on large servers, such as the Earth System Grid Federation \citep{ESGF} for the Climate Model Intercomparison Project \citep{Eyring2016} or the MARS server at the European Center for Medium-range Weather \citep{MARS} forecasts.  As computationally inexpensive ML-based weather forecasting models are increasingly adopted, the volume of simulation output may grow substantially. It may be infeasible to store, share, and transfer huge ensemble simulations (e.g. using open FTP data transfer servers).  In this manuscript, it would be particularly challenging to create a data portal for 2 to 25 petabytes of HENS output.  Depending on the data transfer resources available, transferring the entire 2 petabyte dataset from one location to another may require a prohibitive amount of time and resources.

ML's computational efficiency can lead to a different paradigm of data transfer.  Instead of sharing the simulation output, it may be more practical to share the trained ML model weights\footnote{ML model weights refer to the learned parameters of the architecture: during training, an ML model uses an optimization algorithm to update these weights to minimize the loss function.  At the end of training, the ML model weights can be shared and readily used for inference.} and the initial conditions. This removes the requirement to save the data on a public, high-bandwidth storage location.  By sharing the model weights, other scientific users can regenerate the entire ensemble or a particular subset of members that simulate a particular extreme event of interest.

To ensure reproducibility, we designed HENS based on a random seed.  During inference, this random seed determines the initial condition perturbation, which is created by initially adding spherical random noise to Z500 to start the breeding cycle (see Part I).   During inference, we set the seed for the random number generator.  After the completion of the HENS simulation, we validate that we can regenerate specific ensemble members.  We open-source our ML model weights and inference code for others to regenerate ensemble members of interest. We save the 12 variables listed in Table~\ref{table:variables}.  One could regenerate specific ensemble members to save more variables to assess other atmospheric phenomena during the simulation.   In this way, ML model weights provide a data compression mechanism to avoid storing the data produced from massive ensembles.  However, there are two shortcomings of the seed-based reproducibility.  First, the random-number-generation approach may change in the future as Pytorch and other scientific packages are upgraded.  Second, non-determinism on individual GPUs (e.g. bit-level soft errors, in which the in-memory state of the model could change during the simulation) or in specific frameworks \citep{Vonich2024} pose a challenge for reproducibility.  Although we can reproduce our simulations on our computing environment, bit-for-bit reproducibility is an important direction for future research, especially across different hardware and computing systems.  To assess whether either of these two problems has occurred, checksums of the original ensemble output can be used to ensure that the ensemble regeneration occurred successfully.

\section{Climate and Extreme Statistics}

\subsection{Sampling the Forecast Distribution with Huge Ensembles}

In Part I, we validated a 58-member SFNO-BVMC ensemble using the CRPS, spread-error ratio, threshold-weighted CRPS (twCRPS), reliability diagram, and Receiver Operating Characteristic (ROC) curve.  On these metrics, SFNO-BVMC performs comparably to the Integrated Forecast System (IFS), a leading ensemble numerical weather prediction model based on traditional physics solvers.  These metrics show that SFNO-BVMC has a realistic, reliable, and well-calibrated ensemble spread.  These are prerequisite characteristics for running huge ensembles. If they are not met, the forecast distribution would not provide useful probabilistic information.  Sampling the tails of such a forecast distribution would not yield realistic estimates of internal variability.  In this section, we analyze the sampling properties of the HENS forecasts. 

We explore the large sample behavior of HENS at a 10-day lead time. We present three reasons for this choice of lead time. First, at 10 days, the spread-error ratio has reached approximately 1 for all variables (shown in Figure 8 of Part I), indicating that the ensemble has a reasonable representation of its uncertainty.  Second, 10-day forecasts can be validated against observations because they are within the predictability limit of approximately 14 days.  The 10-day ensemble mean root mean squared error (RMSE) is still lower than the climatology RMSE. Therefore, the HENS trajectories are still dependent on the initial conditions, and they are realistic possible outcomes from the initial conditions.  After the predictability limit is breached, the ensemble distribution should more closely represent the climatological distribution. In this scenario, the model becomes free-running, since its forecast is no longer directly tied to the initial conditions.  At 10 days, we can still directly compare the forecasts to the observations, and we do not have to rely on comparison of the climatological characteristics of the free-running model (e.g. comparison of the probability density functions across space and time).  Third, after 10 days, the ensemble trajectories have diverged due to uncertainty in synoptic-scale atmospheric motion.  This greater ensemble dispersion allows for a more thorough characterization of different future outcomes from the initial time.  


\begin{figure}[t]
\includegraphics[trim={0 0 0 23}, clip, width=0.6\textwidth]{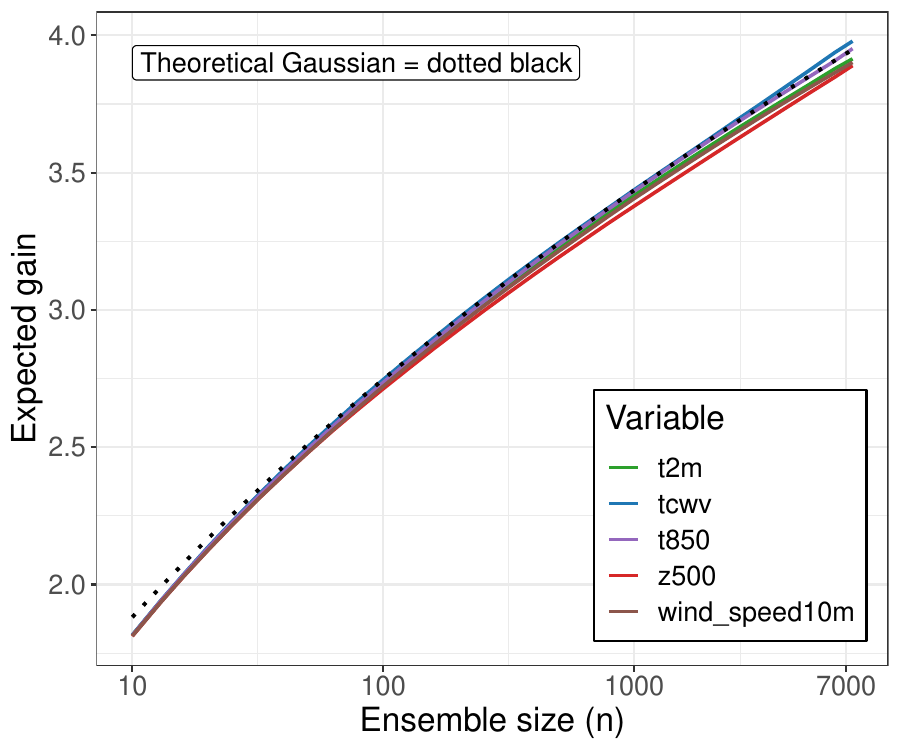}
\caption{\textbf{Information Gain from Huge Ensembles (HENS)}.   Information gain is the maximum number of standard deviations from the mean that are sampled by the ensemble. The mean and standard deviation are calculated from the ensemble distribution itself. Gain is calculated for the ensemble predictions of the global land-mean value of each variable.  For a Gaussian distribution, the theoretical information gain as a function of ensemble size is shown with the dotted black line.  Using the HENS hindcasts from a 7,424-member ensemble initialized each day of boreal summer 2023, the empirical gain for each variable is shown as a function of ensemble size.  Results are shown for a \linebreak {240-hour} lead time (forecast day 10). Note the use of a logarithmic scale on the x-axis.}
\label{fig:information_gain}
\end{figure}
 
To demonstrate the benefit of larger ensembles, we calculate the information gain, $G_n$, for an ensemble of size $n$.  $G_n$~measures the maximum number of standard deviations from the ensemble mean that is sampled by any ensemble member.  Mathematically, this is defined as

\begin{equation} \label{eq:infogain}
G_n = \max_{i=1,\dots,n} \frac{|X_i -  \overline{X}_n|}{S_n}
\end{equation}
where
\[
\overline{X}_n = \frac{1}{n}\sum_{i=1}^nX_i, \hskip5ex S_n = \sqrt{\frac{1}{n-1}\sum_{i=1}^n(X_i-\overline{X}_n)^2}.
\]
Here, $X_i$ is the global land mean value of a given variable for ensemble member $i$, where $i$ goes from 1 to $n$.  Intuitively, $G_n$~measures an ensemble's ability to sample the tails of the forecast distribution.   We assess the expected information gain, i.e., $E[G_n]$, as a function of $n$.

If the tails of the forecast distribution were Gaussian, sampling 4 standard deviation events from the ensemble mean would require approximately 7,000 members.  Section~\ref{app:InfGain} theoretically derives the expected information gain for samples from a Gaussian distribution. The theoretical Gaussian gain is shown by the dotted line in Figure~\ref{fig:information_gain}.  We use this estimate of 7,000 to guide our choice of ensemble size for HENS.   We are constrained by available memory, computational resources, and data movement time from scratch to rotating disk file storage.  Based on these constraints, we determine that we have the computational budget for 7,424 members: 256 members each for 29 trained SFNO checkpoints.  (See Section 2.2 of Part I.) 

We calculate the information gain of HENS $\widehat{G}^{\text{\tiny{HENS}}}_{n}$ in a Monte Carlo sense (here and throughout, ``$\>\widehat{\cdot}\>$'' denotes a statistical estimate).   For a given ensemble size, the information gain is the mean of 2000 bootstrap random samples. For $r = 1, \dots, 2000$ and a given ensemble size $n$, we:
\begin{enumerate}
    \item Calculate $X_i$, the global land mean value of ensemble member $i$ for a given variable,
    \item Randomly sample $n$ values from $\{X_i: i = 1, \dots, 7424\}$, and
    \item Calculate $G_n(r)$ from Equation~\ref{eq:infogain}.
\end{enumerate}
Then $\widehat{G}^{\text{\tiny{HENS}}}_{n} = \frac{1}{2000}\sum_{r=1}^{2000} G_n(r)$. We consider the following variables: 2m temperature, total column water vapor, 850hPa temperature, 500hPa geopotential, and 10m zonal wind.  Due to the nature of the instantaneous atmospheric flow, each ensemble member predicts mean conditions at some locations and extreme conditions at other locations. At a given time, there will likely be extreme conditions occurring somewhere on Earth, simply due to the spatial variation of weather. We do not consider the spatial distribution of extremes, which varies significantly \textit{within} each ensemble member.  Instead, we wish to assess the distribution \textit{across} ensemble members. Therefore, we calculate the information gain on the global land mean values of each ensemble member.  This allows us to assess the ensemble members in aggregate and how far each ensemble member is from the ensemble mean.

Figure~\ref{fig:information_gain} shows the information gain as a function of ensemble size. The 7,424-member ensemble has an information gain of 4.  This means that HENS is large enough to have at least one ensemble member that is 4 standard deviations away from the ensemble mean (on average).  For all global land-means of these variables, the HENS gain closely follows the theoretical Gaussian gain. This result is not completely surprising: averaging over many grid cells implies that a Central Limit Theorem should apply (even though the grid box values are neither independent nor identically distributed), wherein the global land averages behave similarly to a Gaussian random variable.  We use this theoretical Gaussian behavior to inform our choice of ensemble size. With such a large empirical information gain, HENS includes ensemble members that simulate trajectories of low-likelihood events.  To sample ensemble members that are 5 standard deviations away from the mean, $O(10^6)$ ensemble members would be necessary according to the Gaussian approximation.  This ensemble size would require a large compute expenditure and inline ensemble pruning.  Since it would not be practical to save the entire ensemble, it would be necessary to only save the ensemble members that simulate the rarest of extreme events \citep{Webber2019}. 

\begin{figure}[t]
\includegraphics[width=\textwidth]{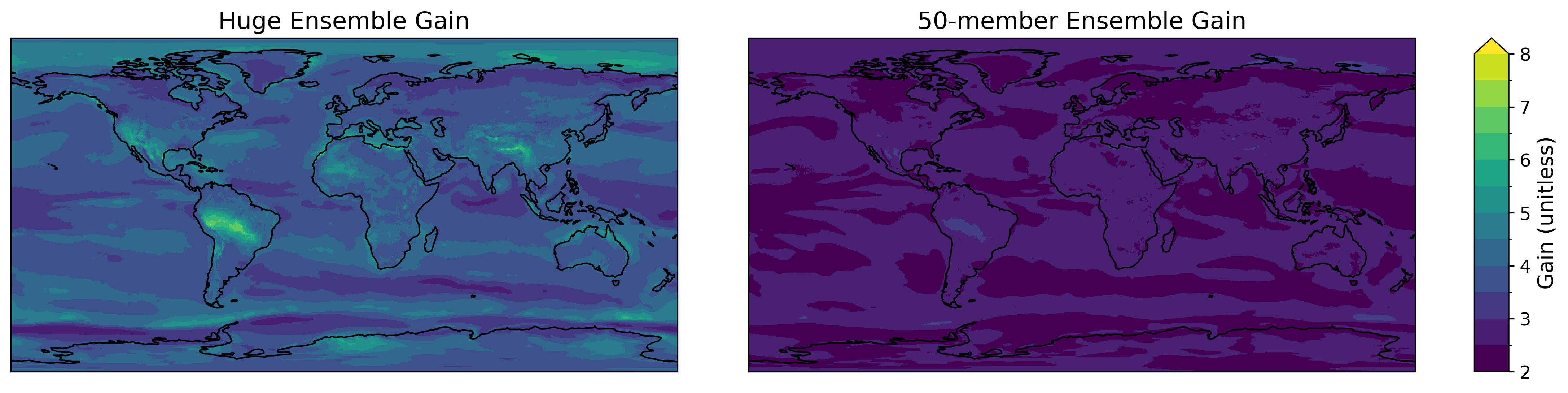}
\caption{\textbf{Information Gain from Huge Ensembles (HENS) at each grid cell}.   Information gain is calculated using the same method as Figure~\ref{fig:information_gain}, but it is calculated at each grid cell, instead of on the global land mean value.  (a) Information gain for huge ensembles (7424 members).  (b) Information gain for 50-member ensembles.  Gain is calculated for 2m temperature at a lead time of 10 days and across all forecasts initialized in summer 2023.}
\label{fig:information_gain_gridcell}
\end{figure}

To complement our analysis of the gain of the global land means, we also assess the gain of huge ensembles at each grid cell.  While the gain of the global-land mean value closely follows the Gaussian approximation, there are significant deviations from Gaussianity at the local level.  In Figure~\ref{fig:information_gain_gridcell}, we visualize the information gain at each grid cell. In the HENS ensemble, some grid cells depart from the expected Gaussian value of approximately 4.  Similarly, in a 50-member ensemble, the grid cells depart from the expected Gaussian value of approximately 2.

\begin{figure}[t]
\includegraphics[trim={0 0 0 27}, clip, width=0.8\textwidth]{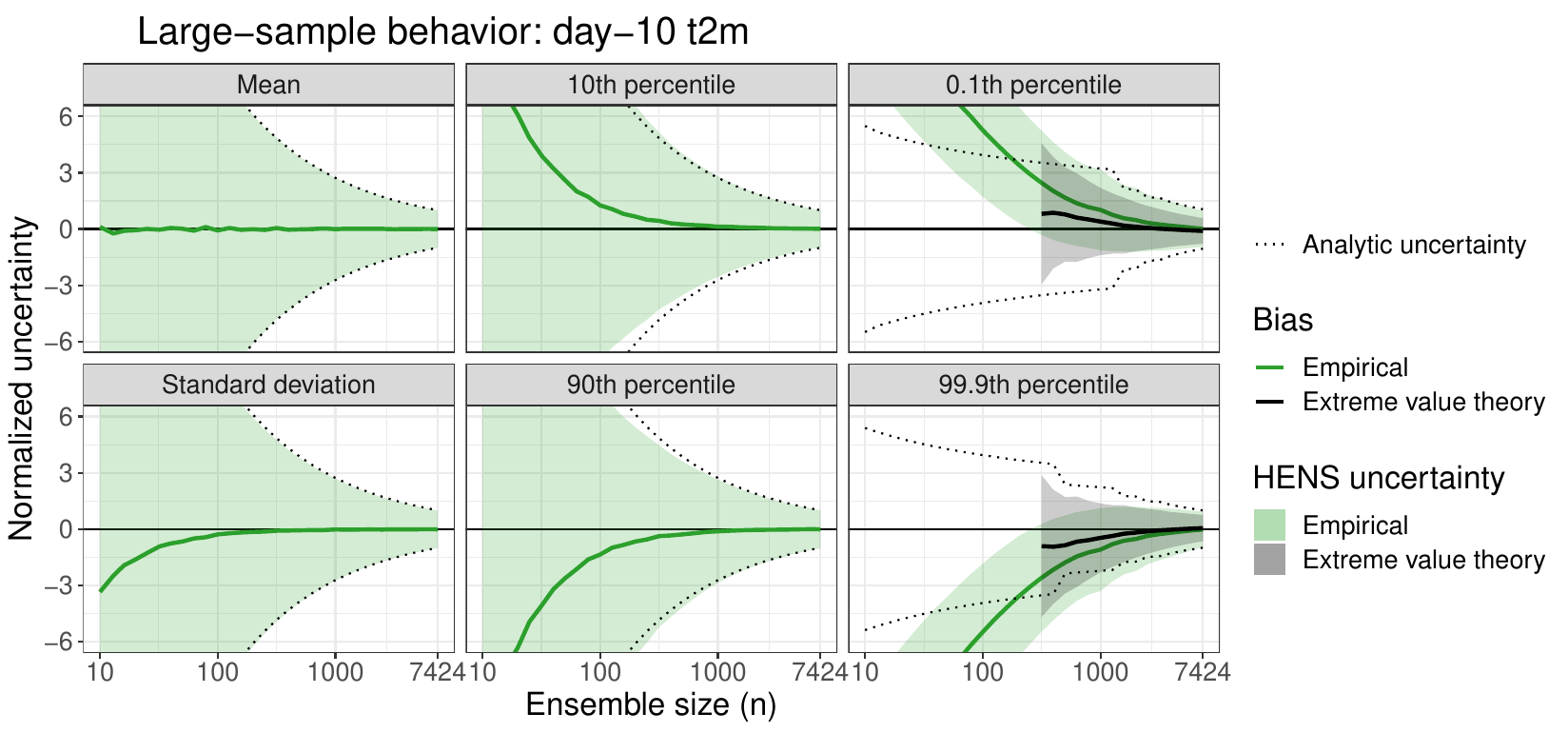}
\caption{\textbf{Large Sample Behavior of Huge Ensembles (HENS)}. The ensemble mean, standard deviation, 0.1$^\text{th}$, 10$^\text{th}$, 90$^\text{th}$, and 99.9$^\text{th}$ percentiles of global land-mean 2m temperature are shown for different ensemble sizes.  For comparison across initial times, all statistics are normalized by the full ensemble standard deviations calculated separately for each forecast initial date.  Statistics are averaged over 92 initial times (one for each day of boreal summer 2023 at 00:00 UTC). The ``true'' statistic is calculated from the full 7,424-member huge ensemble; the solid green line and shading indicate the mean and 95 percent confidence interval, respectively, calculated from bootstrap random samples from the ensemble. Statistics are shown for a 240-hour lead time (forecast day 10). }
\label{fig:large_sample_behavior}
\end{figure}

Using the full HENS distribution, we calculate the ensemble mean, standard deviation, and 0.1$^\text{th}$, 10$^\text{th}$, 90$^\text{th}$, and 99.9$^\text{th}$ percentiles of global land-mean 2m temperature. (Note that the 0.1$^\text{th}$ and 99.9$^\text{th}$ percentiles represent the 1000-day extreme low and high thresholds, respectively.)  Figure~\ref{fig:large_sample_behavior} assesses the ability of different ensemble sizes to emulate these statistics accurately.  For each ensemble size, we take 2,000 bootstrap random samples from HENS, and the resulting statistics and their uncertainty are shown relative to the corresponding statistic from the full ensemble.   We provide a detailed description of calculating these statistics in Section~\ref{app:EstStats}.  In Section~\ref{app:AnalyticUncertainty} and Figure~\ref{Fig_percentile_unc}, we present Gaussian theory for how the percentiles should change with ensemble size $n$.  We use this theory to calculate the "analytic uncertainty" dotted lines in Figure~\ref{fig:large_sample_behavior}. In Section~\ref{app:EVTPercentile}, we use extreme value theory to calculate theoretical estimates of the extreme percentiles in Figure~\ref{fig:large_sample_behavior}. Additionally, in Section~\ref{app:Aggregation}, we discuss how the normalized uncertainty is calculated.  Because of the seasonal cycle in the ensemble spread across summer 2023, we normalize the statistics by the uncertainty from the full HENS forecast at each initial date. This enables comparison of the sampling characteristics across all forecasts from summer 2023.  

For all statistics, smaller ensemble sizes result in large uncertainties and, in some cases, large biases relative to the full ensemble. Estimates of the HENS mean and standard deviation are unbiased (i.e., the empirical bias is near zero) with as few as 10 members and 200 members, respectively. However, these estimates are associated  with large sampling uncertainty, well exceeding six times the uncertainty of the full ensemble.   Larger ensembles are necessary for unbiased estimation of the 10$^\text{th}$ and 90$^\text{th}$ percentiles, on the order of $n=1000$. For ensembles smaller than 1000, the sign of the bias is notable: estimates of the 10$^\text{th}$ and 90$^\text{th}$ percentile are too large and too small, respectively, highlighting the under-sampling of even these moderate percentiles for smaller ensembles.  
Even with 1000 members, there is still sampling uncertainty associated with moderate percentiles, nearly three times that of the full ensemble, and larger ensembles are necessary to reduce this uncertainty. 
For the most extreme percentiles, representing  1000-day events, nearly the full ensemble is needed to obtain empirical estimates that are unbiased.    
In Section~\ref{app:EVTPercentile}, we present extreme value methods to calculate the 0.1$^\text{st}$ and 99.9$^\text{th}$ percentiles.  Compared to directly calculating the percentiles, extreme value theory leads to better estimation of the extremes (Figure~\ref{fig:large_sample_behavior}): estimates are unbiased for ensembles as small as 3000, which is a significant improvement relative to empirical estimates which require $n=7000$. However, as with the 10$^\text{th}$ and 90$^\text{th}$ percentiles, for smaller ensembles both the empirical and extreme value theory estimates are not extreme enough, again illustrating that smaller ensembles do not properly sample the extreme tails of the distribution. 

Across all statistics considered here, larger ensembles lead to significantly more confident estimation of each property of the full 7,424-member HENS ensemble.  These results are robust across different lead times (Figure~\ref{appendexfig:t2m_large_sample_behavior} and for a different variable (see the sampling behavior of 10m wind speed in Figure~\ref{appendexfig:wind_speed10m_large_sample_behavior}).  This is a key value-add of HENS: it enables confident characterization of both the mean and extreme statistics of the forecast distribution, and it quantifies the uncertainty associated with smaller ensemble sizes.  Using this information, the users can select a desired ensemble size based on their use case and an acceptable level of uncertainty. 

For these analyses (and all future analyses in this manuscript), we note that we do not assume the ensemble distribution is Gaussian.  Gaussianity was an emergent property of the global land means, so for these spatial averages, Gaussian theory served as a good estimate of the analytic uncertainty of the ensemble statistics and the information gain.  However, at each grid cell, there are significant deviations from Gaussianity (Figure~\ref{fig:information_gain_gridcell}).  For both global and local forecasts, HENS can robustly sample farther into the tail of the forecast distribution, compared to a 50-member ensemble.  In the next sections, we empirically assess the utility of huge ensembles for weather forecasts and for calculating extreme statistics, and we do not make assumptions about the shape of the distributions.

We present a demonstration of the ensemble forecasts at a specific location during a heatwave in the USA Midwest. Kansas City, Missouri, USA had a significant heat-humidity event on August 23, 2023 at 18:00 UTC. According to ERA5, the 2m~temperature and dewpoint reached 307 K and 298 K, respectively.  To consider the combined effect of both temperature and humidity at the surface, we calculate the heat index introduced by \citet{Lu2022}, which updates the heat index presented in \citet{Steadman1979} to account for particularly hot and humid events.  At this time, the heat index in Kansas City was 316 K (43 C).

\begin{figure}[t]
\includegraphics[trim={0 0 0 29}, clip, width=\textwidth]{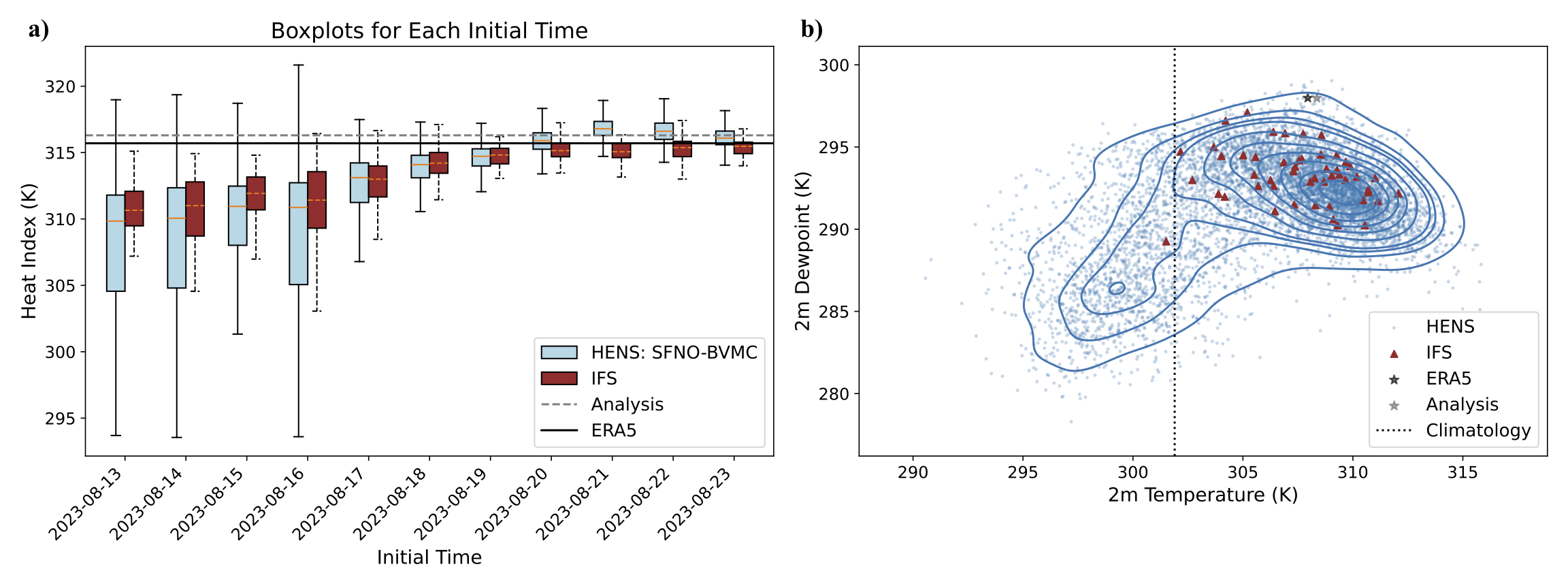}
\caption{\textbf{Demonstration of using Huge Ensembles for heatwave forecasts in Kansas City, Missouri, USA}. (a) Box plot of ensemble forecast of heat index, as a function of initial time.  Blue denotes HENS forecasts and red denotes IFS forecasts.  Range of box and whisker plots indicates the farthest data points within 1.5x the interquartile range. (b) 2D density plot for 10-day forecasts of 2m dewpoint and 2m air temperature.  The outermost contour interval is the 95$^\text{th}$ percentile kernel density estimate of the ensemble distribution.  Contour intervals decrease at intervals of 10 percent.  Blue dots indicate forecasts of individual HENS members; magenta triangles indicate forecasts from IFS ensemble members; the black star is ERA5 (the verification dataset for HENS); and the gray star is operational analysis (the verification dataset of IFS). The dashed line is the climatological average temperature at this location.}
\label{fig:demo_kc}
\end{figure}

Figure~\ref{fig:demo_kc}a shows the HENS forecasts of the heat index, as a function of initial time.  Despite the significantly larger size of HENS, its ensemble spread becomes narrower with lead time.   This indicates that HENS is not overpredicting extreme values at all lead times; its forecasts still have coherent spread as a function of lead time. At a 10-day lead time, both IFS and HENS predicted a warmer than average temperature.  However, the verification air temperature and dewpoint temperature lie at the tails of both ensembles' forecast distributions (Figure~\ref{fig:demo_kc}b). At a ten-day lead time,  none of the IFS members successfully capture the magnitude of both 2m air temperature and 2m dewpoint temperature.  However, HENS does include members that capture the simultaneous intensity of both these values (Figure~\ref{fig:demo_kc}b).  With the large sample size from HENS, it would be possible to study the precursors, drivers, and statistics of the observed extreme. Figure~\ref{fig:demo_kc} uses the same method to visualize large ensembles as \citet{Li2024}.

HENS also enables exploration of counterfactual realities. Some HENS members projected that the heatwave could have been warmer and drier (Figure~\ref{fig:demo_kc}b).    Other HENS members indicate that the heatwave may have not occurred altogether, as they predict 2m temperatures near or below the climatological mean.  These different counterfactuals can result in different climate impacts: hot and humid extremes can be particularly challenging for human health, while hot, dry extremes can have adverse impacts on crop yields and create conditions conducive to wildfire spread. IFS does not contain any sampling of the counterfactual hot, dry version of the event beyond 311 K air temperature and 291 K dewpoint.  We hypothesize that HENS can be used to characterize the dynamical drivers and physical processes associated with each outcome.  In particular, HENS could enable a clustering analysis of the drivers and large-scale meteorological patterns that would result in the observed hot, humid heatwave; a hot, dry heatwave; or no heatwave at all.  If HENS is reliable and well-calibrated for a given type of extreme, it can also be used to study the conditional probability (given the initial conditions) of different outcomes on medium-range weather time scales.

\subsection{Sampling the Observed Distribution with Huge Ensembles}\label{sec:climate_stats_observed_distribution}

\begin{figure}[t]
\includegraphics[trim={0 0 0 27}, clip, width=0.5\textwidth]{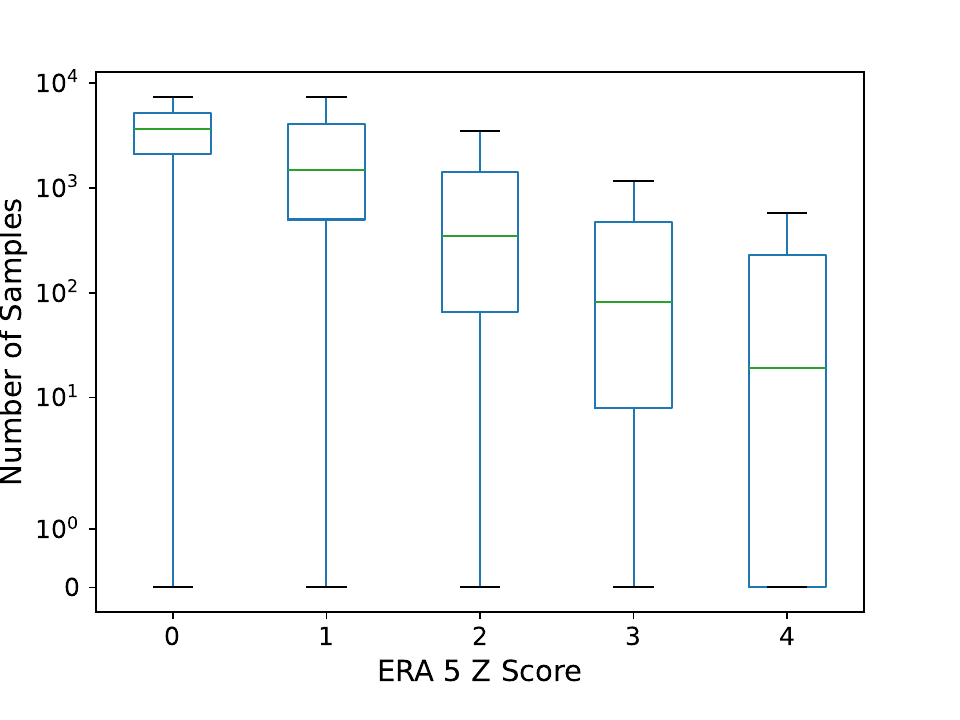}
\caption{\textbf{Number of Ensemble Members for Each ERA5 Event}. From June 1, 2023 to August 31, 2023, the number of ensemble members that have a Z score at least as high as the ERA5 Z score are shown. At each grid cell and time, the Z score  represents how many standard deviations the ERA5 data point is from the mean. The mean and standard deviation are calculated separately for each month and each hour of the day, using the ERA5 climatological periods from 1993 to 2016.  Results are shown for day-10 forecasts (240, 246, 252, and 258 hour lead times) and are averaged over forecasts initialized on each date in June, July, August 2023}
\label{fig:num_ensemble_members}
\end{figure}

In the previous section, we discussed using HENS to study tail events of the \textit{forecast} distribution.  Next, we assess HENS' ability to characterize tail events of the \textit{observed} distribution.  A suitable model for this task must meet two key requirements: (1) it must provide a large sample size, and (2) it must accurately simulate the observed events. 

First, for each grid cell in ERA5, we calculate the climatological mean and standard deviation using a 24-year climatology from 1992-2016.  The climatological mean and standard deviation are calculated for each month for each hour. This is similar to the definition of extreme thresholds used in Part I, and this definition allows the climatology to change for the seasonal and diurnal cycles. (Note that this is distinct from $S_n$ and $\overline{X}$ in Equation~\ref{eq:infogain} above, which use the mean and standard deviation of the ensemble distribution, not the climatological distribution.) 

At each grid cell and time in summer 2023, we convert the ERA5 value into its Z score. Figure~\ref{fig:num_ensemble_members} shows that HENS provides large sample sizes for the ERA5 events that occurred in summer 2023.  In the majority of cases, HENS includes multiple ensemble members that simulate an event that is at least as extreme as the verification value in ERA5. For even the rarest events that are 4 standard deviations away from their climatological mean, HENS includes at least $O(10)$ samples of events with at least that magnitude.  For events that are 2 and 3 standard deviations away, there are hundreds of ensemble members in HENS that meet or exceed the ERA5 value. We note that there are instances where HENS misses an event entirely: these instances correspond to the whiskers of each event having 0 samples.  We quantify this occurrence for heat extremes in Section~\ref{sec:weather_forecasts}.  

\begin{figure}[t]
\includegraphics[width=0.7\textwidth]{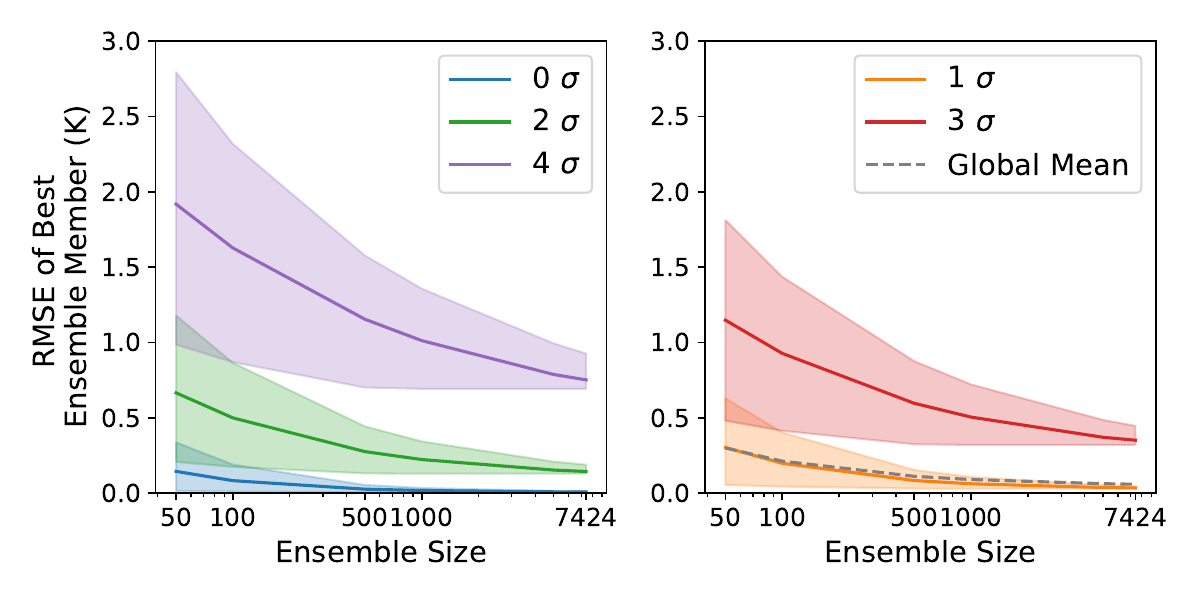}
\caption{\textbf{Skill of the Best Ensemble Member}. For each grid cell, the best ensemble member of a forecast is the ensemble member with the lowest RMSE.  The RMSE of the best ensemble member is shown as a function of ensemble size.  The dashed gray line shows the RMSE of the best ensemble member, averaged over all grid cells and forecasts.  Colored lines show the result for specific $\sigma$ values at certain times, from $0\sigma$ to $4\sigma$ events.  At each location, $\sigma$ represents the number of climatological standard deviations away the ERA5 value is from the climatological mean.  Shaded estimates are the 95 percentile confidence interval calculated from 100 bootstrap random samples at each ensemble size.  All results are for day-10 forecasts (240, 246, 252, and 258 hour lead times) and are averaged over forecasts initialized on each date in June, July, August 2023. Results are spread across 2 panels for better visualization of the shaded estimates associated with each $\sigma$ event.}
\label{fig:rmse_best_member}
\end{figure}

The second requirement for using HENS to study rare observed events is that at least some ensemble members should accurately simulate the true event. With these accurate members, it would be possible to study the event's likelihood and the physical drivers, in comparison to counterfactual outcomes.  To assess the suitability of HENS for these types of analyses, we calculate the RMSE of the best ensemble member.  \cite{Buizza1998} originally introduced this metric for ensembles with up to 32 members.  At each grid cell, they choose the ensemble member with the smallest RMSE across all members.  Using multiple initialized forecasts and lead times, they quantify the RMSE of this best member.  

This metric is actively used in the study of extreme weather events to identify possible drivers. In an operational ensemble weather forecast, \citet{Mo2022} identify the members that had the most accurate forecasts of the 2021 Pacific Northwest Heatwave.  These best members correctly forecast the extent and inland location of a warm-season atmospheric river, which served as a source of latent heat for the heatwave.  \citet{Leach2024} also  examine the ensemble member that is nearest to the observed temperatures. They show that ensemble members that predicted warmer temperatures were associated with low cloud cover and a high geopotential height anomaly. For other types of extreme weather, \citet{Millin2022} identify the ensemble members that most accurately simulated a cold air outbreak, and they show that these members correctly forecast two wave breaks, which were dynamical drivers of the event.   With HENS, these types of analyses can be conducted at scale.  

HENS reduces the RMSE of the best ensemble member; with larger ensembles, this metric systematically decreases by up to 50\% (Figure~\ref{fig:rmse_best_member}). Compared to smaller ensembles, HENS more thoroughly covers the space of possible future outcomes.  In this metric, the HENS improvement is greater for more extreme observed events, indicating the benefits of using HENS to study LLHIs.  These results are robust across lead time (Figure~\ref{appendexfig:rmse_best_member_lead4}) and for a different variable (Figure~\ref{appendexfig:rmse_best_member_wind_speed10m_lead4}).   This opens the door for future analysis to determine why these members correctly forecast the event and to validate whether they forecast the right value for the right reasons.  

We note that the minimum ensemble RMSE metric is not useful in making weather forecasts; while huge ensembles help ensure that at least one member will reasonably match the observations, there is no way of knowing ahead of time which member that will be.


\section{Validating Huge Ensemble Weather Forecasts}\label{sec:weather_forecasts}

\subsection{Metrics based on the entire distribution and metrics based on the conditional distribution}

\begin{figure}[t]

\centering
\makebox[\textwidth][c]{%
        \includegraphics[width=\textwidth]{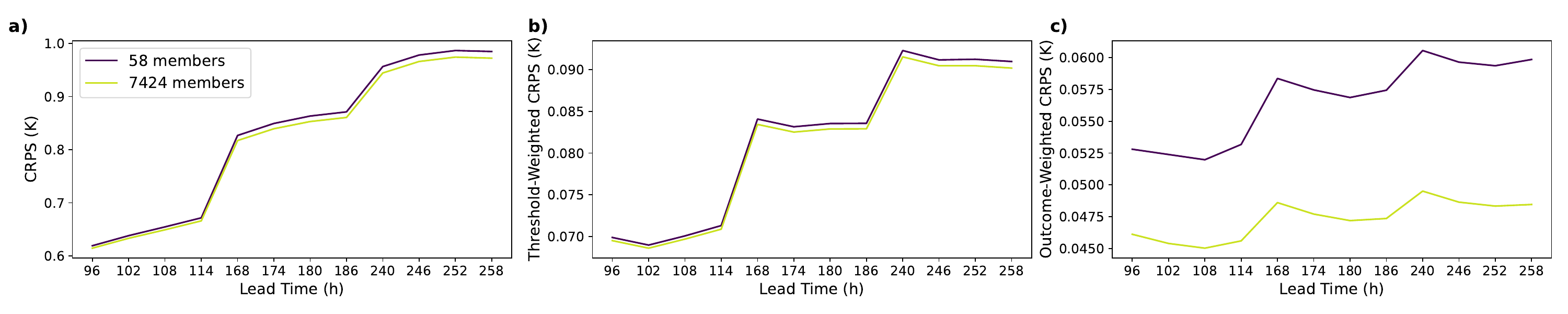}
    }
\caption{\textbf{HENS Continuous Ranked Probability (CRPS) Scores}.  (a) shows overall CRPS scores.  (b) shows threshold-weighted CRPS scores for 95th percentile events.  (c) shows outcome-weighted CRPS scores for 95th percentile events. All scores are calculated at lead times of 4 days, 7 days, and 10 days.  The scores are the global mean and are averaged over 92 initial dates, one for each day in June, July, August 2023.}
\label{fig:crps_comparison}
\end{figure}

Next, we discuss opportunities to improve extreme weather forecasts with larger ensembles.  We compare the 58-member ensemble in Part I to the 7,424-member HENS.  Even though HENS includes more detailed sampling of the forecast distribution, its overall CRPS and threshold-weighted CRPS (twCRPS) scores do not change significantly.  We provide a theoretical basis and a case study to understand the invariance of these two scores with ensemble size. 
These scores are calculated using the entire forecast distribution, but HENS is beneficial for sampling the conditional distribution, given that the forecast is greater than a climatological percentile.  We quantify this benefit through the outcome-weighted CRPS (owCRPS). 


%


On the CRPS and twCRPS scores, HENS performs slightly better than the 58-member ensemble, but the improvement is less than approximately five percent (Figure~\ref{fig:crps_comparison}a and b). We note that we are not using versions of CRPS that are debiased with respect to ensemble size \citep{Zamo2017}.  The twCRPS is calculated analogously to CRPS except all values of the ensemble forecast below a pre-specified threshold are converted to the threshold itself \citep{Allen2023}. As a function of ensemble size, these scores appear to have largely converged with 58 members.  

CRPS and twCRPS compare the CDF of the ensemble forecast to the verification value (see Equations 1 and 3 of Part I). The CDFs are constructed from all members of the ensemble.   The Dvoretzky–Kiefer–Wolfowitz (DKW) inequality \citep{Massart_1990} quantifies the maximum difference between a true population CDF and an empirical CDF constructed from $n$ samples.  Here, the HENS forecast is the population CDF, and the 58-member ensemble is the sample empirical CDF.  The difference between these two CDFs scales with a factor of $1/\sqrt{n}$ (Appendix Section~\ref{app:SamplingECDF}).  With this scaling factor, a 58-member CDF closely approximates a 7,424-member CDF, so we expect the corresponding CRPS and twCRPS to be very similar.  Based on this argument, we also expect the Extreme Forecast Index, which is also calculated from a CDF based on the entire ensemble (see equation 2 of Part I), to be very similar in HENS and the 58-member ensemble. 

However, HENS has an advantage in resolving the conditional distribution, given that the ensemble is above a climatological threshold.  This conditional distribution is created by truncating the ensemble forecast distribution at the climatological 95th percentile.  When the CRPS is calculated using the conditional distribution, it is referred to as the outcome-weighted CRPS (owCRPS). Mathematically, owCRPS is

\begin{equation} \label{eqn:owcrps}
    \text{owCRPS(F, y, w)} = w(y)CRPS(F_w, y)
\end{equation}

where  $w$ is the weighting function, $y$ is the verification value from ERA5, and $F_w$ is the CDF of the conditional ensemble distribution:

\begin{equation}
    F_w = \frac{E[1\{X \leq x\}w(X)]}{E[w(X)]}
\end{equation}

where $X$ is a random variable from the ensemble CDF.  We use the weighing function $w(y) = 1\{y > t\}$, based on a threshold~$t$, which is the 99th percentile of 24-year ERA5 climatology as the threshold.

\begin{figure}[t]
\includegraphics[width=\textwidth]{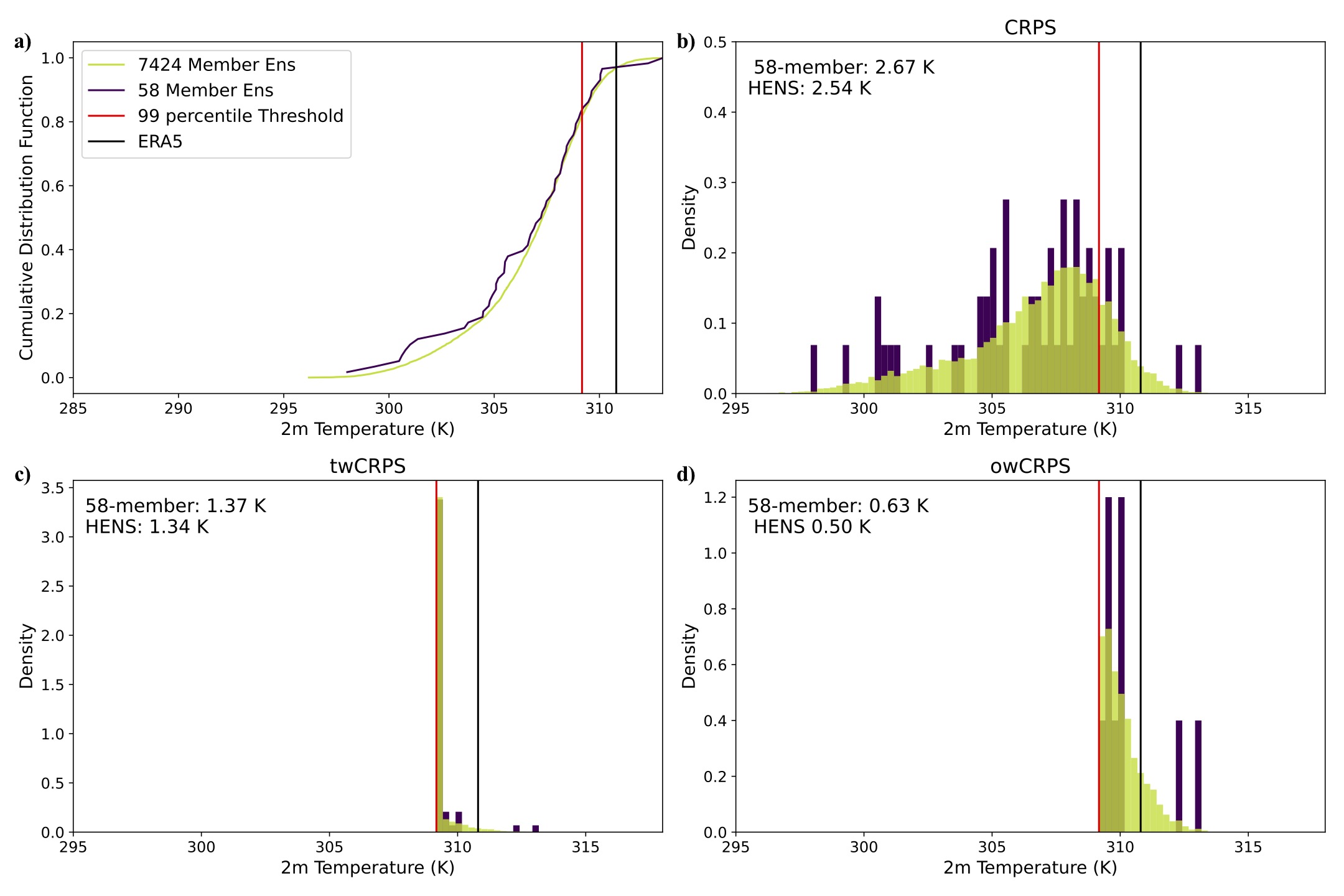}
\caption{\textbf{Visualization of CRPS, threshold-weighted CRPS, and outcome-weighted CRPS}.  Sample forecasts from HENS and the 58-member SFNO-BVMC (which is described in Part 1).  Forecasts are for 2m air temperature near Shreveport, Louisiana, USA, initialized on August 13, 2023 00:00 UTC and valid on August 23, 2023 00:00 UTC. (a) shows the CDFs of the HENS forecast and the 58-member forecast.  (b) shows the forecast distributions from these two ensembles used to calculate the overall CRPS.  (c) shows the distributions used to calculate the threshold-weighted CRPS.  The threshold-weighted CRPS is calculated using the forecast distribution, transformed such that all members below the 95th percentile are set to have the value of the 95th percentile itself. (d) shows the forecast distributions used to calculate the outcome-weighted CRPS.  The outcome-weighted CRPS is calculated using the conditional forecast distribution, conditioned on the forecasts being greater than the 95th percentile threshold. For CRPS, twCRPS, and owCRPS, the numbers in the top left correspond to the scores achieved by the 58-member ensemble and HENS for this forecast. Note the different y axes for (b), (c), and (d) due to the use of different distributions in the CRPS, twCRPS, and owCRPS.}
\label{fig:crps_demo}
\end{figure}

In Figure~\ref{fig:crps_demo}a, we show a case study of the full ensembles' CDFs at Shreveport, Louisiana, USA.  Due to its smaller size, there is more noise in the 58-member CDF compared to the HENS CDF.   In Figures~\ref{fig:crps_demo}b, c, and d, we compare the distributions that are used to calculate CRPS, twCRPS, and owCRPS.  These three scores use the raw ensemble distribution, the transformed ensemble distribution, and the conditional distribution, respectively. twCRPS results in a point mass being placed at the threshold (Figure~\ref{fig:crps_demo}c), since all ensemble members below the threshold are converted to having a value at the threshold itself. owCRPS is created from the conditional distribution.  Figures~\ref{fig:crps_demo}b, c, and d mirror the schematic in Figure 1 of \citet{Allen2023}, which demonstrates calculating CRPS, twCRPS, and owCRPS. 

In this case study, HENS shows small improvements in CRPS and twCRPS, which use the entire ensemble distribution, but it shows a large improvement in the owCRPS, which uses the conditional distribution.  In Figure~\ref{fig:crps_demo}d, the 58-member ensemble had 10 members above the threshold, and HENS had 1340 members above the threshold.  Increasing the ensemble size from 10 to 1340 members yields a greater sampling improvement than increasing the ensemble size from 58 to 7424 members.  In this case study, 10 members are not enough to adequately characterize the tail of the forecast distribution (Figure~\ref{fig:crps_demo}d).  Because of the significant sampling uncertainty, the small ensemble has a different conditional distribution than HENS's conditional distribution.  HENS reduces the owCRPS from 0.63 K to 0.5 K (a relative reduction of about 20\%). Unlike CRPS and twCRPS, the number of ensemble members used in the owCRPS varies for each forecast at each grid cell: it depends on how many members are above the threshold.  

 

The owCRPS is calculated only when the extreme event actually occurs, so it is not a statistically proper scoring rule. Proper scoring rules are minimized when the forecast distribution matches the distribution from which the observation is drawn \citep{Gneiting2007}.  They cannot be hedged by overpredicting extremes (see Part 1 for a discussion on the relationship between CRPS and proper scoring rules for extreme forecasts).  Since the owCRPS is only calculated when extremes actually occur, a forecast that overpredicts extremes could falsely appear reliable \citep{Lerch2017}.  This is the essence of the forecaster's dilemma (see Part I).  However, HENS does not appear to be overpredicting extremes and hedging its scores because it has a comparable CRPS, twCRPS, reliability, and spread-error ratio as the 58-member ensemble.  If HENS were overpredicting extreme weather, then these scores would degrade. In Figure~\ref{appendexfig:hens_vs_small_reliability}, Figure~\ref{appendexfig:hens_vs_small_spread_error_separate}, Figure~\ref{appendexfig:hens_vs_small_spread_error}, we also show that HENS has comparable reliability, spread, and error as the 58-member ensemble.

Using all forecasts initialized during summer 2023 (Figure~\ref{fig:crps_comparison}c), the HENS owCRPS is approximately 20\% better than the \linebreak 58-member ensemble at lead times of 4, 7, and 10 days.  Based on Figure~\ref{fig:crps_comparison}b, HENS moderately improves the ability to sample the bulk distribution (e.g. through the reduced noise in Figure~\ref{fig:crps_demo}), and this could result in the small improvements to CRPS and twCRPS.  In many cases, a 58-member ensemble is adequate to represent the bulk distribution.  Indeed, ensembles of approximately 50 members have been responsible for the tremendous skill of existing ensemble weather forecasts.  To characterize the ensemble distribution conditioned on exceeding a threshold, only a subset of members can contribute to the conditional CDF. In this scenario, HENS provides a significant advantage over traditional ensemble sizes.

\subsection{Confidence Intervals of Extreme Forecasts}

\citet{Leutbecher2018} discuss the effect of sampling uncertainty on ensemble forecasts.  This sampling uncertainty comes from the fact that there is a finite ensemble being used to approximate an underlying forecast distribution. We assess this sampling uncertainty in the context of extreme weather forecasts.

One way to generate an extreme forecast is to binarize the ensemble at a given threshold.  For instance, if 10 out of 58~members predict a climatological 99th percentile temperature event, then the ensemble forecasts a 17\% probability of extreme temperature.  In this probabilistic prediction, there is uncertainty induced by finite sample size. The forecast depends on which 58 members are sampled out of all possible ensemble members.  Bootstrap sampling (sampling with replacement) can be used to approximate the sampling uncertainty. Using the 2.5 and 97.5 percentiles across many bootstrap samples, one can bootstrap a 95 percent confidence interval.  This confidence interval represents the uncertainty due to finite sample size.

In the example from Shreveport in Figure~\ref{fig:crps_demo}, the sampling uncertainty has important implications for making the ensemble forecasts.  The HENS forecast issues an 18\% probability of an extreme weather event, since 1340 out of the 7424 members are above the threshold.  At a sample size of 7424, the HENS 95 percent confidence interval is 17.1\% to 18.9\%; this confidence interval was obtained from taking 2000 bootstrap random samples of the HENS forecast.  On the other hand, with an ensemble size of 58, the probability of an extreme event is 17\%.  However, the 95 percent confidence interval ranges from 8.6\% to 28\%.  This interval is significantly wider, indicating that there is more uncertainty with the 58-member ensemble's forecast. This uncertainty is an order of magnitude larger than the sampling uncertainty from 7,424 members.  

\begin{figure}[t]
\includegraphics[width=0.8\textwidth]{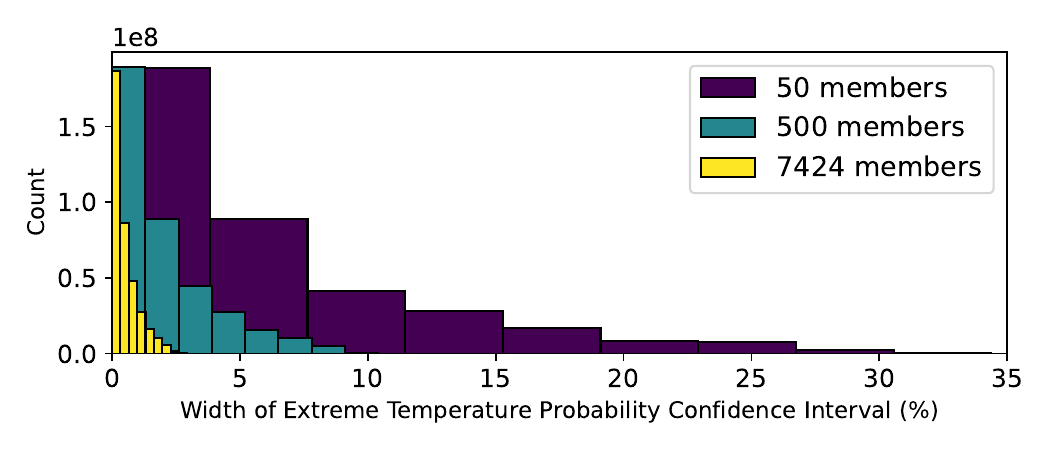}
\caption{\textbf{Effect of Ensemble Size on Forecast Confidence Intervals}. An extreme forecast is issued by categorizing each ensemble member as "extreme" or "not extreme," using the 99th percentile 2m temperature at each location.  The extreme temperature forecast is the percent of ensemble members that are above the threshold.  For each ensemble size, a confidence interval for the extreme forecast is calculated from 100 bootstrap random samples from the ensemble.  For all nonzero extreme forecasts issued in summer 2023 at all locations, the histogram of confidence interval widths is shown for different ensemble sizes. All forecasts have a lead time of 10 days.  On the $y$-axis, the counts are multiplied by a factor of $10^8$, since the histograms are calculated over all grid cells and 92 initial times.}
\label{fig:confidence_interval_width}
\end{figure}

In summer 2023, across all 10-day nonzero extreme forecasts at all grid cells, the confidence intervals are an order of magnitude narrower than those from small ensembles (Figure~\ref{fig:confidence_interval_width}). This result is robust across different lead times: Figure~\ref{appendexfig:confidence_interval_width_two_panel} calculates the same confidence interval width at lead times of 4 days and 7 days.  With less sampling uncertainty for extreme forecasts, there could be more informed disaster readiness and more targeted plans. However, even with narrower confidence intervals, we note that it is still of course possible to have uncertain forecasts.  For instance, if half the ensemble members predict extreme weather, the chance of extreme would still be 50\%.  This probability is rooted in initial condition and model uncertainty. HENS does not alleviate these uncertainties; it reduces the sampling uncertainty of the forecast distribution.  Therefore, with HENS, the 95 percent confidence interval around the 50\% estimate would be narrower. 

\subsection{Missed Events in HENS and IFS}

The outlier statistic measures the ability of an ensemble to capture the true verification value \citep{Buizza1998}.  They define this statistic as the proportion of cases in which the verification lies outside the bounds of the ensemble forecast.  Using HENS, we calculate this statistic as a function of ensemble size. For each ensemble size, we take 100 bootstrap random samples from HENS. At each grid cell, if the ERA5 value lies within ninety-five percent of the bootstrapped ensembles, then the ensemble size is deemed satisfactory for capturing the event.  Otherwise, the ERA5 value is classified as an outlier.  

\begin{figure}[t]
\includegraphics[width=0.4\textwidth]{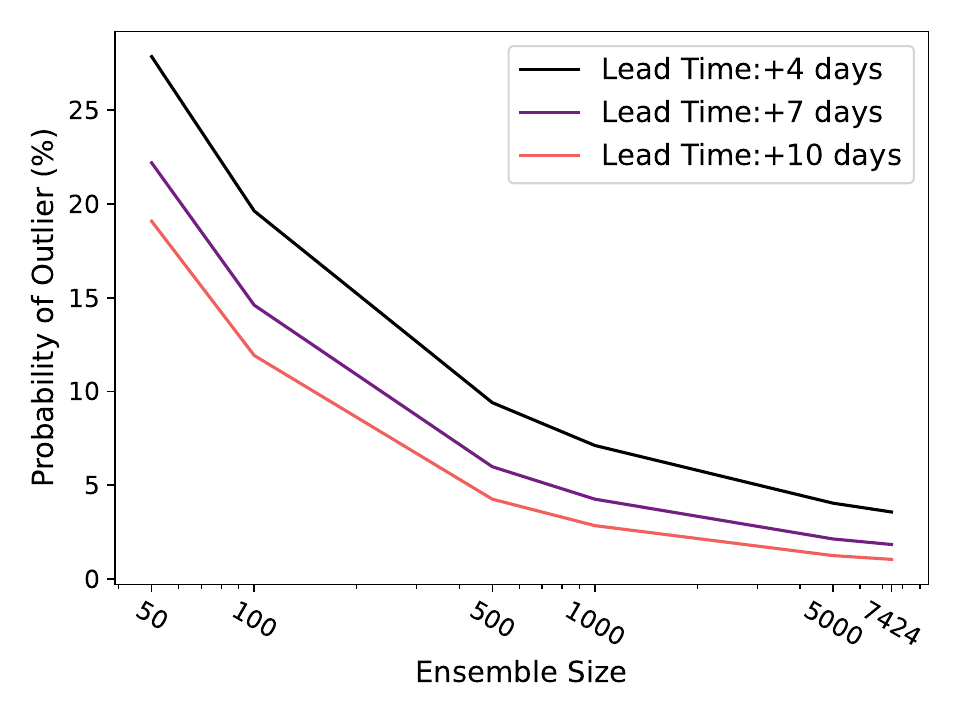}
\caption{\textbf{Outlier Statistic}. An outlier occurs when the ERA5 value lies outside the ensemble range in ninety-five percent of bootstrap ensemble samples (samples from the ensemble with replacement).  The outlier statistic is the proportion of the globe that is covered by outliers.  For different ensemble sizes,  the outlier statistic is calculated using all forecasts initialized in summer 2023 for lead times of 4 days, 7 days, and 10 days.}
\label{fig:outlier_statistic}
\end{figure}

\begin{figure}[t]
\includegraphics[width=0.7\textwidth]{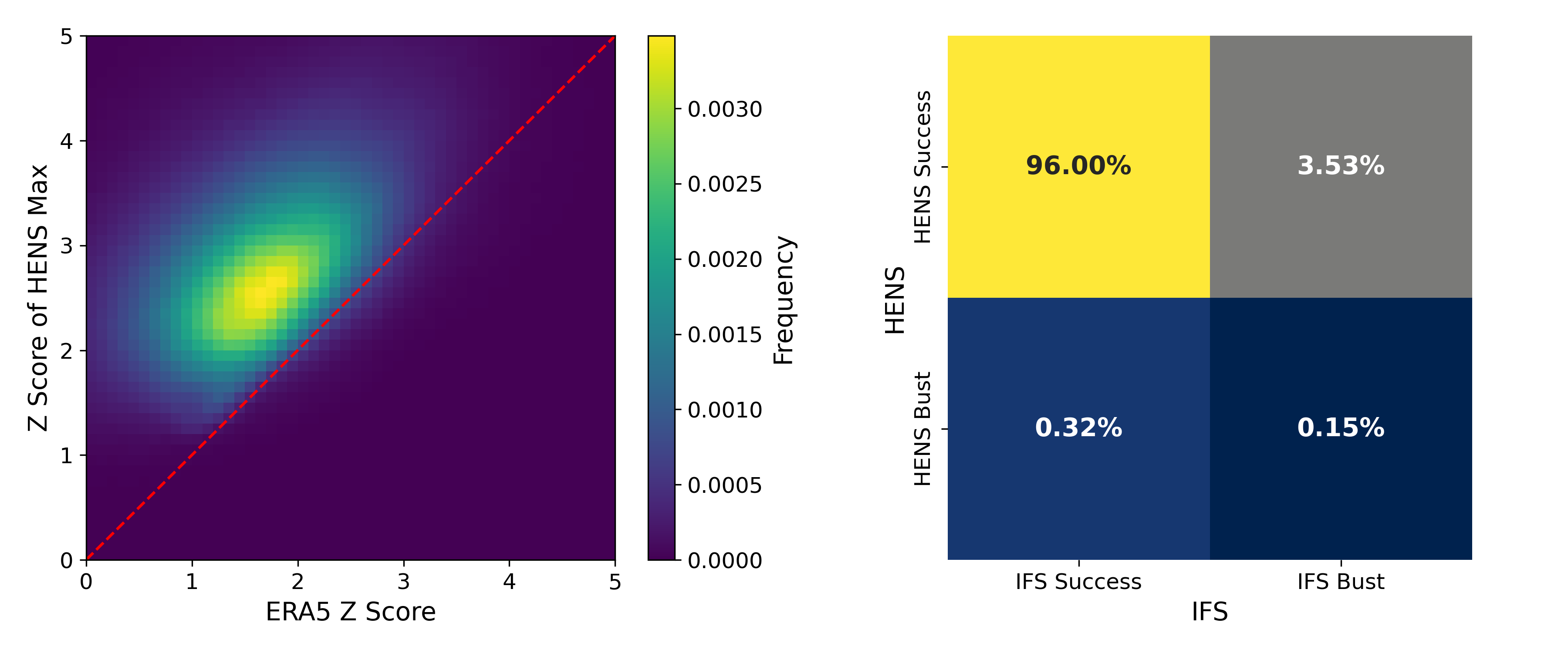}
\caption{\textbf{IFS and HENS Comparison of Missed Warm 2m Temperature Events}.  An IFS bust occurs when the operational analysis is greater than the IFS ensemble max.  A HENS bust occurs when ERA5 is greater than the HENS max. (Left) 2D Histogram of IFS busts.  At times and locations of IFS busts, the histogram shows the corresponding Z Score of the HENS maximum and the ERA5 Z Score.  Z scores are calculated for each grid cell using the ERA5 climatological mean and standard deviation. (Right) Confusion matrix showing proportion of IFS busts, IFS successes, HENS busts, and HENS successes.}
\label{fig:zscore_hensmax}
\end{figure}

Figure~\ref{fig:outlier_statistic} shows larger ensembles reduce the probability of an outlier.  At a lead time of 10 days, the 7,424 member ensemble is satisfactory for representing almost 99\% of the ERA5 values. The reduction in the outlier statistic is robust across lead times of 4 days, 7 days, and 10 days.   The 10-day forecasts have a systematically lower outlier statistic than 7-day and 4-day forecasts because they have the best spread-error ratio (see Part 1).  By taking 100 bootstrap random samples, we modify the calculation of the original statistic presented in \citet{Buizza1998}. Their original statistic does not use bootstrap samples of the ensemble, so it does not account for sampling uncertainty.  Our modification to account for sampling uncertainty tends to increase the outlier statistic. This modification requires 95\% of all possible bootstrapped ensembles to capture the event, rather than just one ensemble.  For a more direct comparison with the \citet{Buizza1998} method, we next calculate the proportion of cases in which the verification dataset is greater than the ensemble max (Figure~\ref{fig:zscore_hensmax}). This is analogous to the warm side of their outlier statistic.  In Figure~\ref{fig:zscore_hensmax}, we do not consider ensemble sampling uncertainty, as in Figure~\ref{fig:outlier_statistic}.  We only analyze warm outliers because summer 2023 was the hottest summer on record, and we focus on the ability of HENS to represent extreme heatwaves.

Figure~\ref{fig:zscore_hensmax} compares the abilities of IFS and HENS to include warm events within their ensembles.  For the warm side of the distribution (Z scores > 0), we calculate the latitude-weighted proportion of cases in which the operational analysis exceeds the IFS ensemble max.  We label these instances "IFS busts."  For those IFS busts, we assess whether the HENS maximum exceeds the ERA5 value.  Figure~\ref{appendexfig:hens_max_ifs_miss_walkthrough} shows a demo of the calculation presented for one forecast at one initial time.  

Across all of summer 2023, Figure~\ref{fig:zscore_hensmax}a shows that the bulk of the distribution lies above the 1-to-1 line, indicating that the HENS maximum successfully captures the heat extremes that IFS missed.  Figure~\ref{fig:zscore_hensmax}a only assesses the performance of HENS during instances of IFS busts. Figure~\ref{fig:zscore_hensmax}b characterizes the relative occurrences of IFS busts and HENS busts using a confusion matrix.  For the majority of cases (96\%), both HENS and IFS capture the event:  most of the time, they both have an ensemble member that is at least as large as the verification.  However, in 3.5\% of cases in summer 2023, IFS did not have a member with a sufficiently large 2m temperature value.  For the vast majority of these IFS busts, HENS did capture the true event.  We note that there is a small portion of cases (0.32\%) where HENS missed an event that IFS successfully captured.  This is the topic of further research and could be due to cases where the HENS ensemble mean is biased.

As shown in Figure~\ref{fig:zscore_hensmax}, many of the IFS busts occur around events that have a Z score of approximately 2.  For these events and other even rarer events, IFS cannot successfully sample the extremes.  HENS can capture these events, yet it still maintains its CRPS (Figure~\ref{fig:crps_comparison}), reliability (Figure~\ref{appendexfig:hens_vs_small_reliability}), ensemble mean RMSE (Figure~\ref{appendexfig:hens_vs_small_spread_error_separate}), ensemble spread (Figure~\ref{appendexfig:hens_vs_small_spread_error_separate}), and spread-error ratio (Figure~\ref{appendexfig:hens_vs_small_spread_error}) of the corresponding 58-member SFNO-BVMC ensemble.

\section{Discussion and Conclusion}

In total, HENS includes $10,245,120$ simulated days, or $28,050$ simulated years (7424 ensemble members $\times$ 15 day rollout per ensemble member $\times$ 92 forecast initialization days).  With such a large sample size, we demonstrate the value-add of HENS: sampling events that are 4 standard deviations away from the ensemble forecast mean, accurately estimating the tails of the ensemble, elevating the skill of the best member, and reducing the likelihood of outliers.  Based on its superior owCRPS scores and its narrow confidence intervals for extreme weather forecasts, HENS samples the tails of the forecast and observed distributions.

HENS can reduce uncertainty in extreme weather forecasts.  Future research is necessary to associate these improved scores (e.g., in the outlier statistic or owCRPS) with an economic value. Based on end users' specific use cases, cost-loss models are guided by economic principles. They can be used to quantify the benefits of improved forecasts and reduced uncertainty \citep{Wilks1995, Palmer2002}.   This analysis would enable a detailed consideration of whether huge ensembles are appropriate and necessary for a given stakeholder.

We evaluate our HENS simulations on the medium-range timescale, up to the predictability limit of approximately 14 days.  On these time scales, it is possible to directly compare the forecast with observations.  The suite of NWP metrics provides insights into the realism of the ensemble forecast. If the ensemble members are exchangeable with each other and with the observations, the spread-error ratio (presented in Part I) should be close to 1.  We show that SFNO-BVMC has a spread-error ratio that is close to 1, especially by a lead time of 10 days. Similarly, HENS has competitive scores on CRPS and twCRPS, which evaluate probabilistic forecasts against the true observed trajectories from ERA5.  Additionally, Figure~\ref{fig:zscore_hensmax}a shows that increasing the ensemble size allowed HENS to capture real events that were missed in the IFS ensemble.  With its larger ensemble size, HENS reduced the number of instances where the verification dataset was greater than the ensemble max (Figure~\ref{fig:zscore_hensmax}b).   Increasing the ensemble size increased the skill of the best ensemble member (Figure~\ref{fig:rmse_best_member}). By increasing the sample size, the ensemble better simulates real extreme events, as measured through owCRPS. To further validate whether the simulations realistically represent the dynamics of the Earth's atmosphere, future research can validate the HENS predictions on idealized test cases \citep{Hakim2024, Mahesh2023, Mamalakis2022}.

A fundamental constraint with simulating LLHIs is that these events are rare by definition, and there are limited observational samples with which to benchmark ensemble forecasting systems. With machine learning, this challenge is further complicated, since forty years of observations are reserved for training.  Here, we demonstrate that huge ensembles of ML-based forecasting systems offer promising results for summer 2023. Future research is necessary to validate these ensemble systems on more LLHIs. In particular, the climate community can invest computational resources in creating large ensembles of physics-based simulations with high horizontal, vertical, and temporal resolution.  These simulations would extend ML and LLHI science in multiple ways.  In perfect model experiments, they can be used as additional out-of-sample simulations with which to validate ML weather prediction models.  Alternatively, these simulations can be used to train the ML emulators, and all years of the observational record can be used as an out-of-sample validation set.

HENS is not a replacement for traditional numerical weather prediction.  Due to computational costs and energy requirements, it is not feasible to scale traditional ensemble weather forecasts to 7,424 members.  HENS is a computationally efficient way to inflate the ensemble size to study and forecast extreme weather events at the tail of the forecast distribution.  For issuing operational weather forecasts, a combination of existing methods and larger ML-based ensembles can be used.  Additionally, HENS relies on ERA5 reanalysis as its training dataset, so it still needs ensemble data assimilation and numerical models for its forecasts.

An important future direction is to consider the effect of ensemble size in tandem with horizontal resolution.  All our results here are based on 0.25-degree horizontal resolution, which is the resolution provided by ERA5.  A key tradeoff in climate science is whether the compute budget is better spent on larger ensembles or finer resolution \citep{Schneider2024}.  Because of their minimal computational cost, SFNO enables analysis of both options.  Future work is necessary to train ML ensembles to emulate kilometer-scale climate datasets (e.g. SCREAM \citep{Caldwell2021}).  This emulation can guide decisions on the tradeoff between ensemble size and horizontal resolution.

To analyze simulations with large data volumes, an important technical frontier is the inline calculation of diagnostics.  Traditionally, the simulation output is saved to disk during the forecast rollout, and then the saved data is loaded into a separate offline diagnostic analysis pipeline. On the other hand, inline diagnostics could be calculated during the ML ensemble generation itself.  Offline diagnostic calculation is feasible for smaller ensembles that produce less data, but for HENS, it poses significant challenges due to I/O and disk storage space limits. Inline diagnostics could help alleviate the I/O and memory challenges associated with analyzing petabytes of simulation output. In particular, for climate analysis applications at scale, I/O can serve as the primary constraint for model analysis.  In this manuscript, we save the entire simulation output and calculate our diagnostics offline, after the simulation is complete.  We discuss the challenges and post-processing necessary for this in Section~\ref{app:postprocessing}.  



Due to their cheap computational cost, ML-based forecasts present an opportunity to study the dynamics and statistics of extreme events.  This opens the door for robust characterization of the drivers of extreme events, such as the heatwave drivers listed in \cite{Domeisen2022}. With large dataset sizes, it becomes more feasible to apply causal analysis frameworks \citep{Runge2019} to identify these drivers. For studying the statistics and drivers of rare events, a larger sample size is crucial. As a database of simulated weather events, HENS samples the tails of weather forecast distributions and is a new resource for analyzing extreme weather.








\codedataavailability{The code, datasets, and models are all stored at \url{https://doi.org/10.5061/dryad.2rbnzs80n}, via DataDryad.  The code is integrated with Zenodo at the prior DOI. We include the code to train SFNO, conduct ensemble inference with bred vectors and multiple checkpoints, and scoring and analysis code.  We also open-source the model weights of the trained SFNO.  See the README of the DOI for information on how to use the codebase and for the permissive license associated with the code and data.  The code is available via the Lawrence Berkeley Lab BSD variant license, and the data is available with a CC0 license. For convenience, the webpage of our project is https://github.com/ankurmahesh/earth2mip-fork.

} 



\appendix

\section{References for Ensembles Listed in Figure 1}\label{app:ensemble_size_reference}

\begin{enumerate}
    \item \citet{Weyn2019}
    \item \citet{Scher2021}
    \item \citet{Weyn2021}
    \item \citet{Pathak2022}
    \item \citet{Hu2023}
    \item \citet{Bi2023}
    \item \citet{Kochkov2023}
    \item \citet{Zhong2024}
    \item \citet{Price2023}
    \item \citet{BaoMedina2024}
    \item \citet{Guan2024}
    \item \citet{Li2024}.  (They also include a demo of a 16,384 member ensemble at one location and lead time.)
    \item \citet{Frame2008}
    \item \citet{Hazeleger2010}
    \item \citet{jeffrey2013australia}
    \item \citet{Rodgers2015}
    \item \citet{Kay2015}
    \item \citet{Sanderson2015}
    \item \citet{KirchmeierYoung2017}
    \item \citet{Sun2018}
    \item \citet{Maher2019}
    \item \citet{Kelder2022}
    \item \citet{Gessner2021}
    \item \citet{Leach2022}
    \item \citet{Craig2022}
    \item \citet{Miranda2023}
    \item \citet{Fischer2023}
    \item \citet{Ye2024}

\end{enumerate}

\section{Post-processing Data to Improve Technical Analysis of the Huge Ensemble}\label{app:postprocessing}

Analysis of petabyte-scale data volumes is challenging and nontrivial. A key challenge relates to the storage of the ensemble in memory.  For certain analysis below, we perform reductions on the ensemble dimension: for each initial time, lead time, latitude, and longitude, we calculate the ensemble mean, maximum, minimum, standard deviation, 99th percentile, bootstrap random sample, or CRPS.  However, for a given initial time, HENS generates one file per ensemble member.  This creates the following dimension ordering of the data in its storage in memory: (ensemble, lead time, latitude, longitude).  For a given initial time, lead time, latitude, and longitude, the ensemble members are stored in separate locations in memory, not in a contiguous chunk.  With our analysis workflow, this is suboptimal for performing ensemble reductions.  It leads to prohibitive I/O read times to load each ensemble member and rearrange the data in memory such that the ensemble dimension can be reduced.  Additionally, loading in the ensemble members in parallel and communicating via the Message Passing Interface (MPI) requires a significant number of concurrent processes and has prohibitive memory demands.

To analyze the ensemble, we post-process 2m temperature and the heat index to be stored in contiguous chunks in memory.   Our post-processing amounts to a matrix transpose.  During the simulation, the data is stored with the following array order: (ensemble, lead time, latitude, longitude).  We transpose this to (lead time, latitude, ensemble, longitude).  Through this transpose, for a given lead time and latitude, all 7424 ensemble members at all 1440 longitudes can be analyzed in a contiguous 43~MB chunk of memory.  This one-time post-processing enables faster analysis workflows that reduce the ensemble dimension; this is because all the ensemble members are stored together in memory.  For more optimal I/O, we use file striping on NERSC's scratch system.  The post-processed file is split such that multiple servers can read the file in parallel.  We perform all ensemble analysis using mpi4py (MPI for Python) and h5py. These tools scale well for dataset sizes of this magnitude, and they are well-optimized for high-performance computing centers and their file systems.

\section{Effect of Ensemble Size on Reliability Diagrams and Spread Error Ratio}

\begin{figure}[t]
\includegraphics[width=0.5\textwidth]{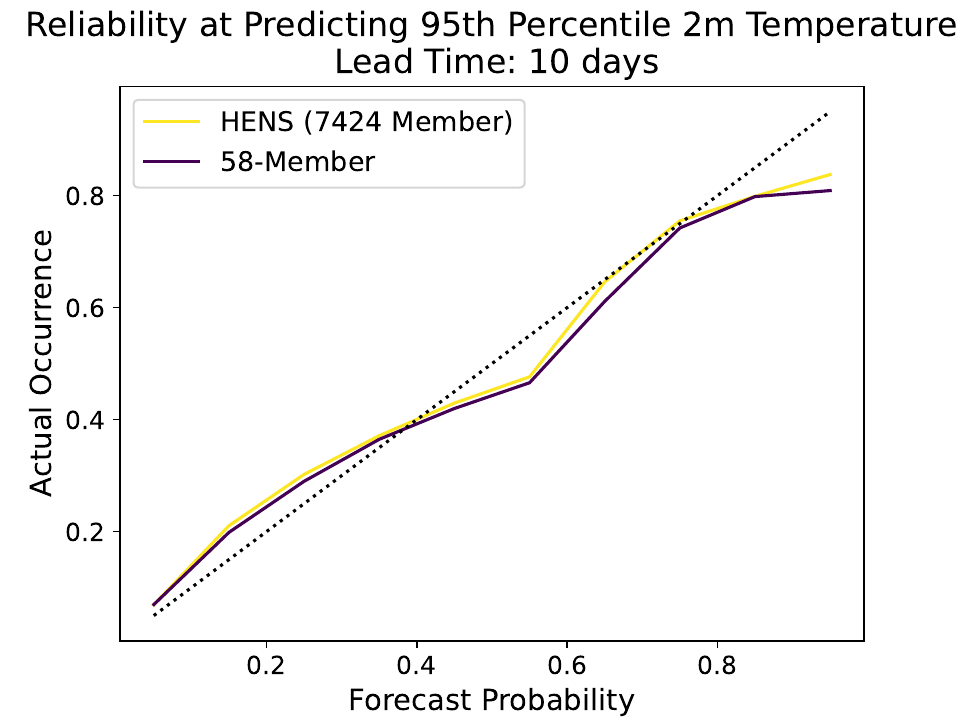}
\caption{\textbf{Effect of Ensemble Size on Reliability Diagram}. The reliability diagram is compared for a 58-member ensemble (purple) and a huge ensemble (yellow).  Results are shown for predicting the 95th percentile 2m temperature, where the 95th percentile is calculated from a 24-year climatology. 
 See Part I for more details. Results are shown for a 10-day lead time for all forecasts in summer 2023. }
\label{appendexfig:hens_vs_small_reliability}
\end{figure}

\begin{figure}[t]
\includegraphics[width=0.5\textwidth]{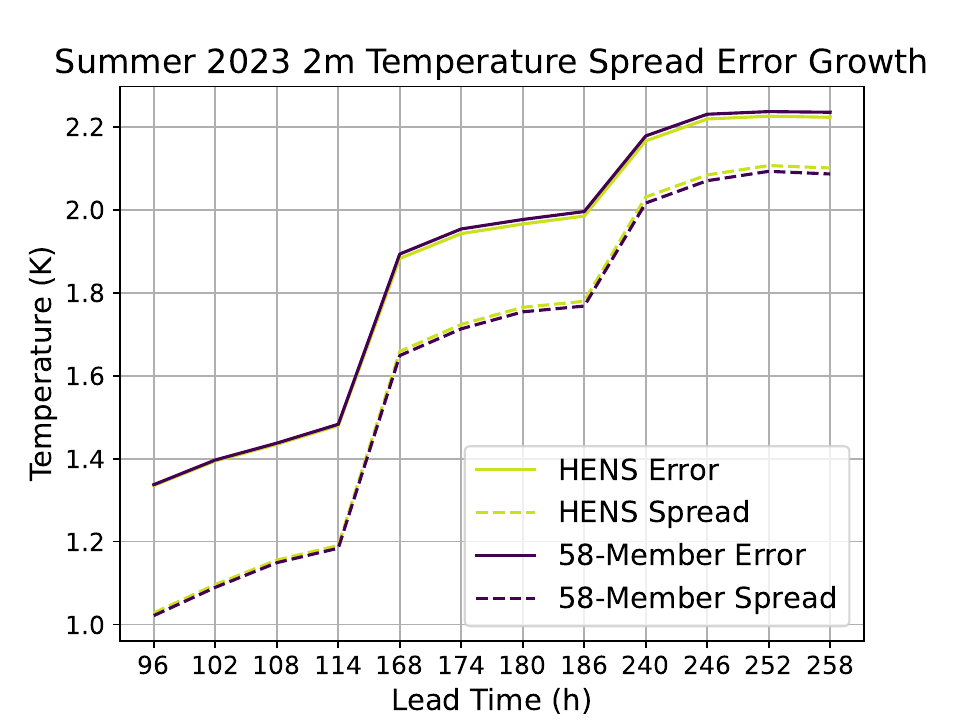}
\caption{\textbf{Effect of Ensemble Size on Ensemble Mean RMSE and Ensemble Spread}. The ensemble spread and the ensemble mean RMSE is compared for a 58-member ensemble (purple) and a huge ensemble (yellow). Results are shown for all forecasts initialized during summer 2023. Note that results are only shown for lead times of 4 days, 7 days, and 10 days.}
\label{appendexfig:hens_vs_small_spread_error_separate}
\end{figure}

\begin{figure}[t]
\includegraphics[width=0.5\textwidth]{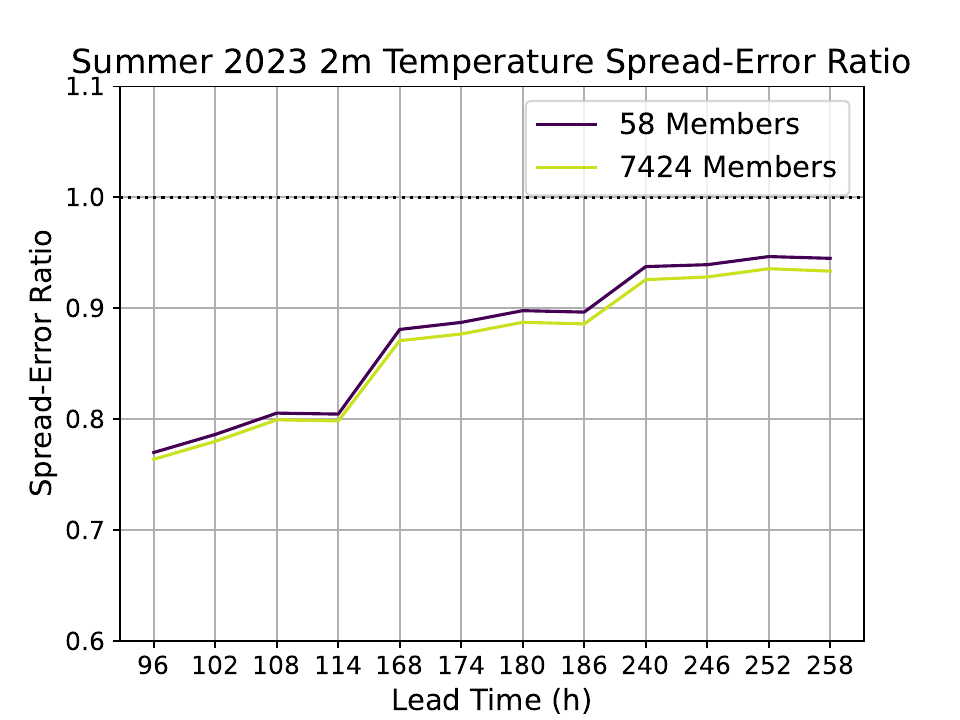}
\caption{\textbf{Effect of Ensemble Size on Spread-Error Ratio}. The spread-error ratio is compared for a 58-member ensemble (purple) and a huge ensemble (yellow). Results are shown for all forecasts initialized during summer 2023. Note that results are only shown for lead times of 4 days, 7 days, and 10 days.}
\label{appendexfig:hens_vs_small_spread_error}
\end{figure}

\section{Effect of ensemble size on confidence intervals}

\begin{figure}[t]
\includegraphics[width=\textwidth]{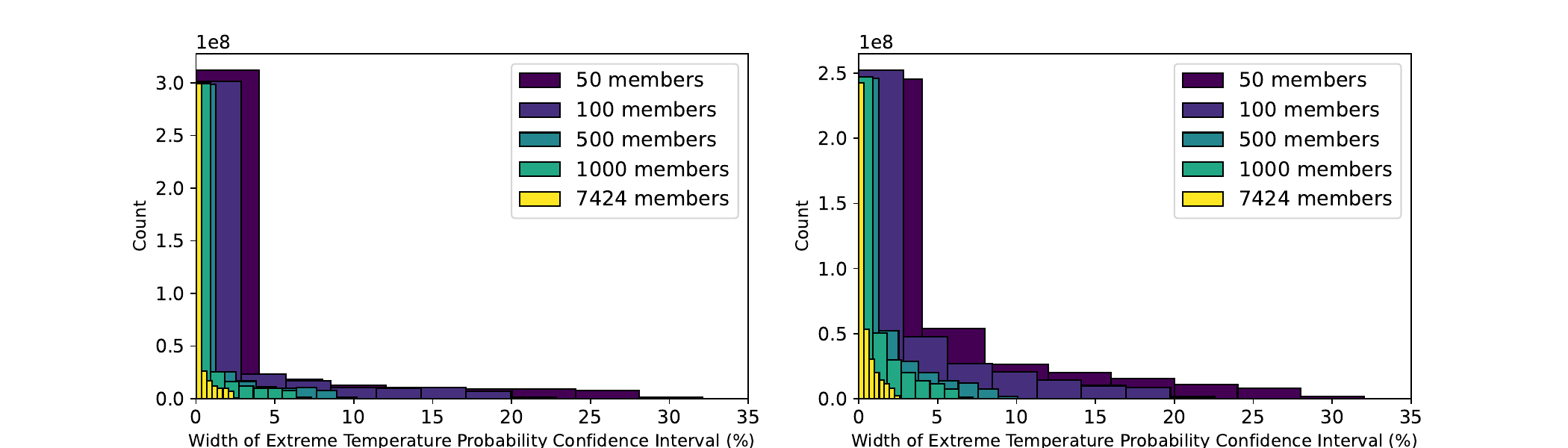}
\caption{Same as Figure~\ref{fig:confidence_interval_width}, but for 4 day lead times (left) and 7 day lead times (right).}
\label{appendexfig:confidence_interval_width_two_panel}
\end{figure}

In Figure~\ref{appendexfig:confidence_interval_width_two_panel}, the effect of ensemble size on confidence intervals is calculated at lead times of 4 days (96-114 hours) and 7~days (168-186 hours).

\section{Example of IFS Busts and HENS Busts}

\begin{figure}[t]
\includegraphics[width=1\textwidth]{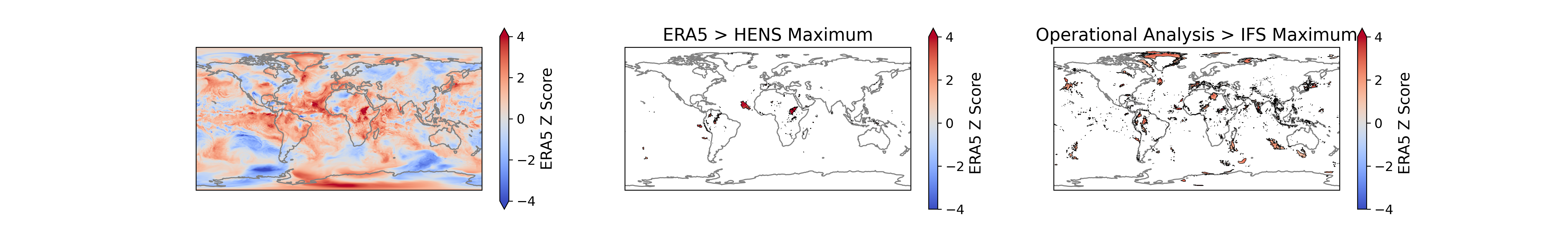}
\caption{\textbf{Walkthrough of HENS Misses and IFS misses}. ERA5 2m temperature on August 23, 2023, converted to Z Scores using the 24-year climatological mean and standard deviation.  Climatological mean and standard deviation calculated for the month (August) and hour (18:00) UTC.  Forecast busts are locations where the verification dataset is greater than the ensemble max. 
Forecast busts are shown for HENS (middle) and IFS (right).  HENS and IFS are initialized on August 13, 2023 at 00:00 UTC and valid on August 23, 2023 at 18:00 UTC time.  Black contours indicate all locations where there is a forecast bust.}
\label{appendexfig:hens_max_ifs_miss_walkthrough}
\end{figure}

Figure~\ref{appendexfig:hens_max_ifs_miss_walkthrough}a shows the ERA5 Z score values at each grid point.  These values are calculated from the climatological mean and standard deviation of 2m temperature in August at 18:00 UTC.  Figure~\ref{appendexfig:hens_max_ifs_miss_walkthrough}b shows the instances where ERA5 was greater than the maximum ensemble member in HENS.  These instances are HENS busts, because HENS was unable to capture the true event.  Figure~\ref{appendexfig:hens_max_ifs_miss_walkthrough}c shows the instances of the operational analysis being greater than the IFS ensemble max, which are analogously busts for IFS.  Figure~\ref{appendexfig:hens_max_ifs_miss_walkthrough}b and c show that HENS has a smaller proportion of cases with a bust.  In Figure~\ref{fig:zscore_hensmax}, we show the aggregate results from performing this calculation over all forecasts from all 92 initial times, initialized in summer of 2023.

\section{Assessing Robustness of HENS Across Lead Times and Variables}

\begin{figure}[t]
\includegraphics[width=0.6\textwidth]{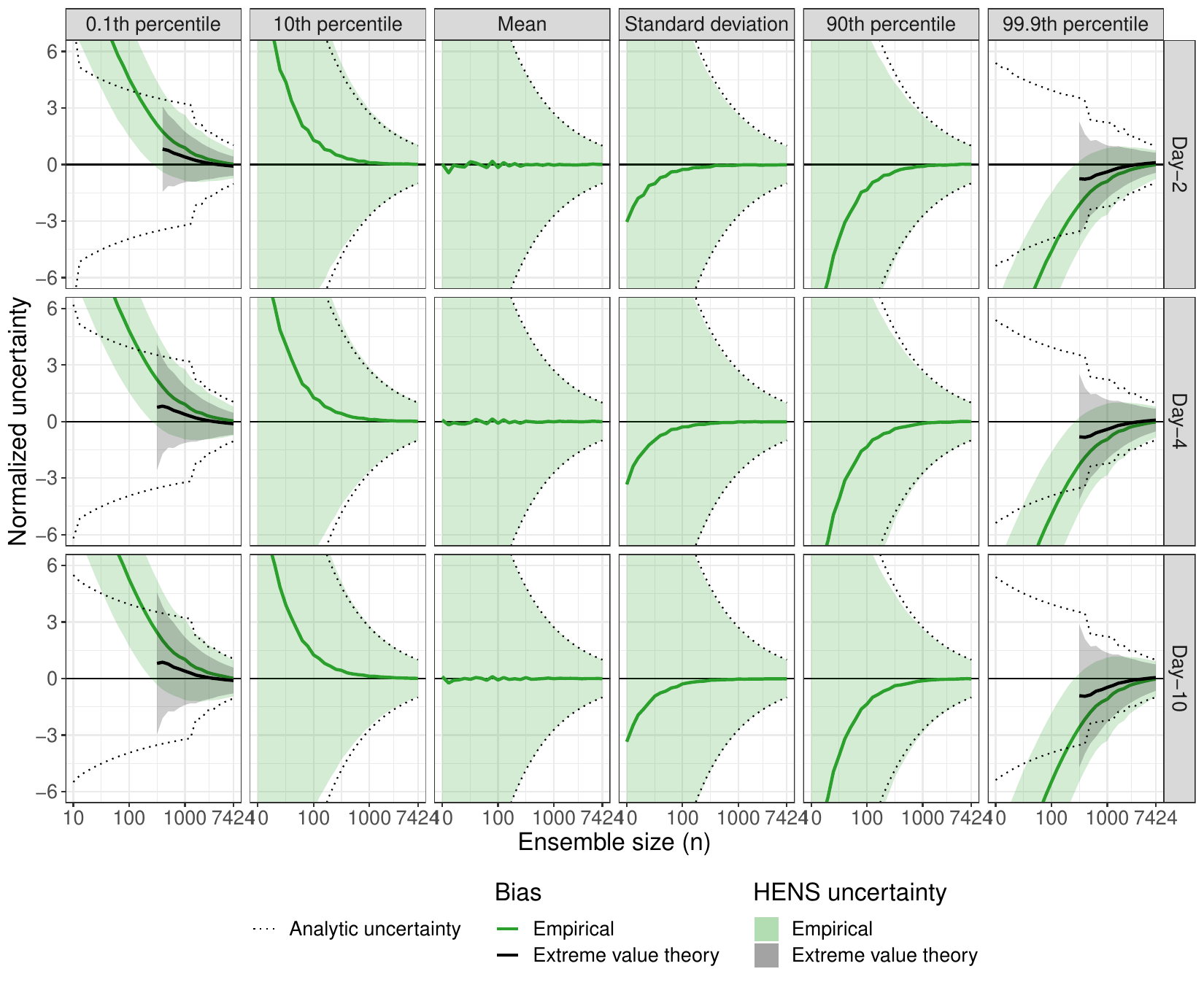}
\caption{\textbf{Large Sample Behavior of Huge Ensembles (HENS)}. This figure is the same as Figure~\ref{fig:large_sample_behavior}, but it compares day-2, day-4, and day-10 forecasts of 2m temperature.}
\label{appendexfig:t2m_large_sample_behavior}
\end{figure}

\begin{figure}[t]
\includegraphics[width=0.6\textwidth]{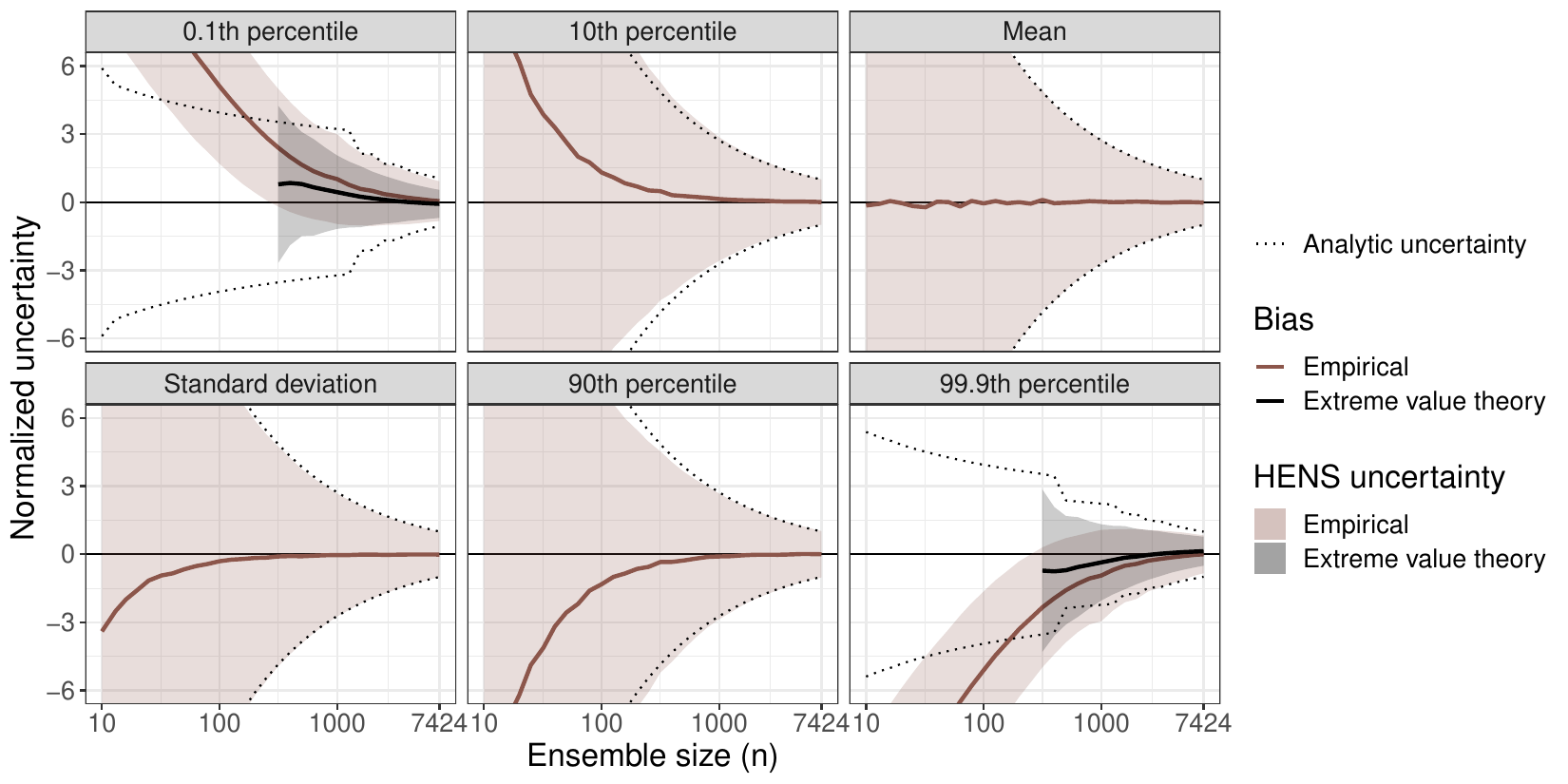}
\caption{\textbf{Large Sample Behavior of Huge Ensembles (HENS)}. This figure is the same as Figure~\ref{fig:large_sample_behavior}, but for 10m wind speed at a lead time of 4 days.}
\label{appendexfig:wind_speed10m_large_sample_behavior}
\end{figure}

\begin{figure}[t]
\includegraphics[width=0.7\textwidth]{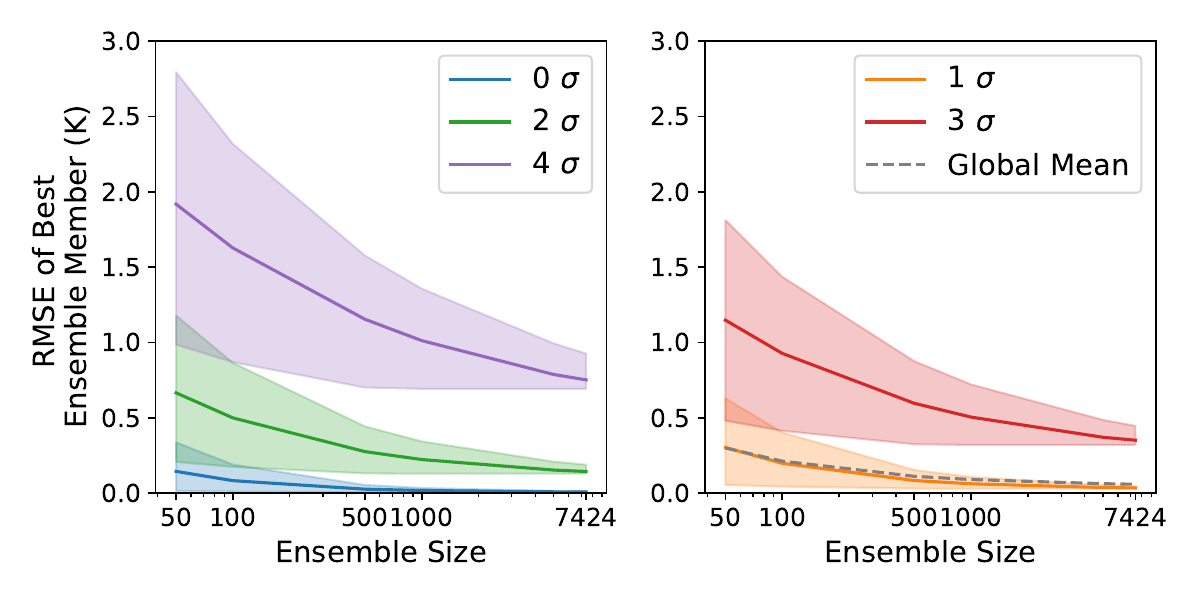}
\caption{\textbf{Skill of the Best Ensemble Member}. This figure is the same as Figure~\ref{fig:rmse_best_member}, but for a lead time of 4 days.}
\label{appendexfig:rmse_best_member_lead4}
\end{figure}

\begin{figure}[t]
\includegraphics[width=0.7\textwidth]{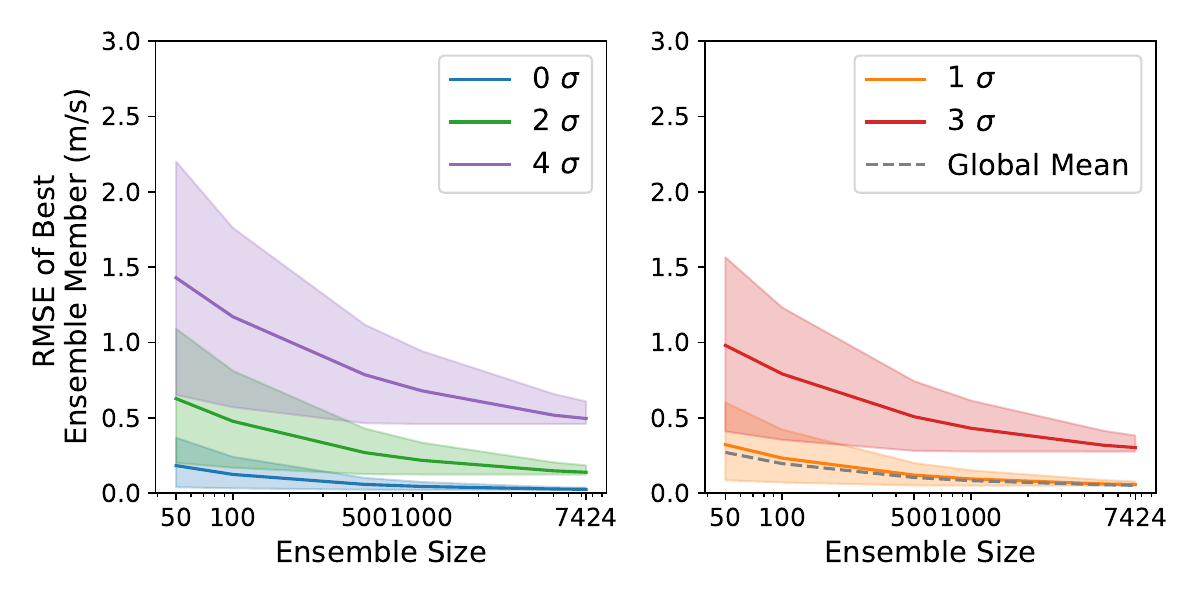}
\caption{\textbf{Skill of the Best Ensemble Member}. This figure is the same as Figure~\ref{fig:rmse_best_member}, but for 10m wind speed at a lead time of 4 days.}
\label{appendexfig:rmse_best_member_wind_speed10m_lead4}
\end{figure}


\section{Statistical properties of the ensemble}

\label{app:StatsEnsemble}

\subsection{Setup and notation}
\label{app:Setup}

\noindent Define a sample of random variables $X_1, \dots, X_N$ that are independent and identically distributed (IID) according to a probability density function $f(x)$. Similarly, define a cumulative distribution function $F(x) = \int_{-\infty}^x f(y) dy$. We later take the $\{ X_i : i = 1, \dots, N\}$ to be the day-10 global land-averages of an output variable from the HENS experiment for a given initialization date, such that $N=7,424$.

\subsection{Exchangeability} \label{sec_exc}\label{app:Exchangeability}

We first assess the extent to which there is a ``signature'' of the model checkpoints in the day-10 forecasts, i.e., are the output of the different model checkpoints  interchangeable? This is commonly (in statistics) referred to as \textit{exchangeability}. One way to assess the relative differences between model checkpoints is to propose a statistical model that compares the between-model (or inter-model) variability and the within-model (or intra-model) variability. If we re-index the HENS output to be $\{X_{ij}: i = 1, \dots, 29; j = 1, \dots, 256 \}$, where $i=1, \dots 29$ indexes the individual model checkpoints and $j=1, \dots, 256$ indexes the ensemble members for each checkpoint (yielding $29\times256=7,424$), the statistical model assumes 
\begin{equation} \label{eq_randomeffects}
X_{ij} \sim N(m_i, \tau^2), \hskip4ex m_i \sim N(m, \sigma^2);
\end{equation}
i.e., each ensemble member arises from a Gaussian distribution with checkpoint-specific mean and variance $\tau^2$ (within-model variance the same for all models), and the mean of each model checkpoint arises from a different Gaussian distribution with an overall mean and variance $\sigma^2$. Here $\sigma^2$ describes the between-model variance. Note that this framework is robust to the specific distribution assumed  in Eq.~\ref{eq_randomeffects} \citep{McCulloch_2011}. The broader assumption is that the specific 256~ensemble members for each model are representative of a potentially much larger pool of draws from that model; furthermore, that the 29 models come from a broader population of models that could have been used. In any case, we are then interested in the so-called ``exchangeability ratio'' $R = \sigma/\tau$ to assess the relative magnitude of the inter- and intra-model spread. 
If $R > 1$, then this suggests the models are more different from one another than the individual ensemble members from a given model are; if $R < 1$, this suggests the within-model spread dominates the between-model spread. 
In other words, $R<1$ implies exchangeability. We can also calculate a 95\% confidence interval on this ratio \citep[Appendix A]{Longmate_2023}.

\begin{figure}[t]
\includegraphics[width=0.4\textwidth]{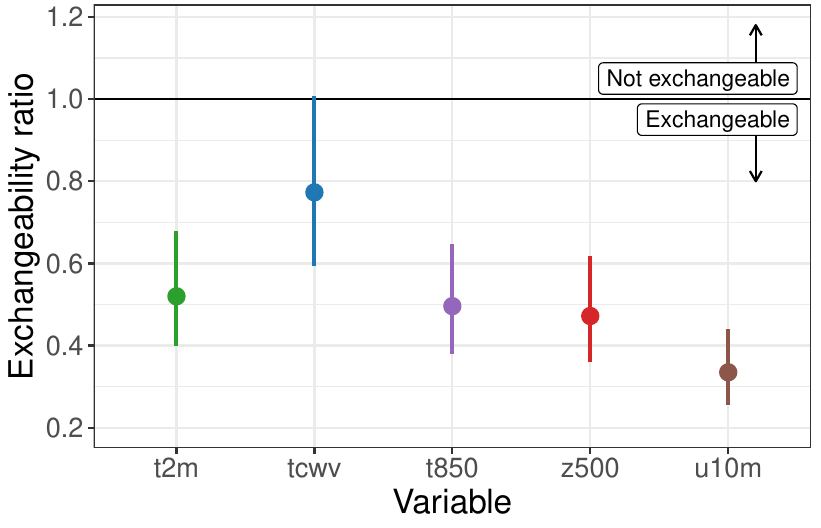}
\caption{\textbf{Exchangeability Ratio}. Using the global land mean values of each ensemble, the ensemble spread between SFNO checkpoints is compared to the ensemble spread within one checkpoint.  The ratio of the inter- and intra- model spread is shown for 2m temperature (t2m), total column water vapor (tcwv), 850hPa temperature (t850), 500hPa geopotential height (z500), and 10m wind zonal wind (u10m).  Exchangeability ratios are shown for 10-day lead time across forecasts initialized in summer 2023. The vertical lines denote the 95\%~confidence interval for this ratio.}
\label{appendexfig:exchangeability_ratio}
\end{figure}

\subsection{Expected information gain}
\label{app:InfGain}

\noindent Define the \textbf{information gain} for a sample of $n$ random variables to be
\begin{equation} \label{eq_infogain}
G_n = \max_{i=1,\dots,n} \frac{|X_i -  \overline{X}_n|}{S_n}
\end{equation}
where
\[
\overline{X}_n = \frac{1}{n}\sum_{i=1}^nX_i, \hskip5ex S_n = \sqrt{\frac{1}{n-1}\sum_{i=1}^n(X_i-\overline{X}_n)^2}.
\]
Our goal is to assess the expected information gain, i.e., $E[G_n]$, as a function of $n$.

\subsubsection{Special case: Gaussian} 
\label{app:Gaussian}

\begin{figure}[!t]
\begin{center}
\includegraphics[trim={0 0 0 0mm}, clip, width = 0.75\textwidth]{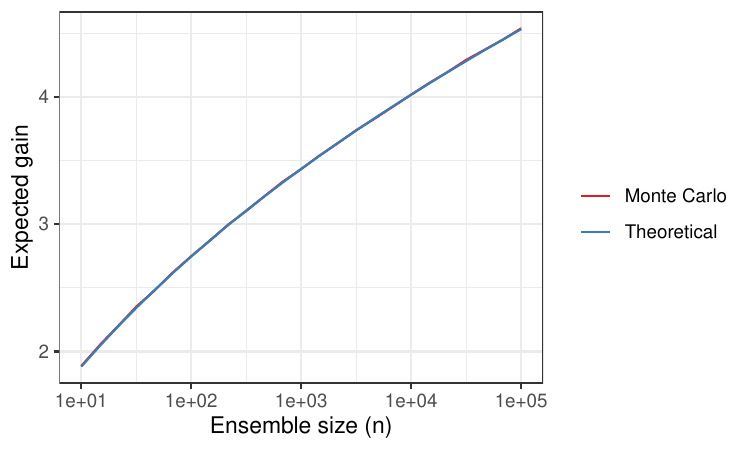}
\caption{Expected gain $E[G_n]$ as a function of the ensemble size $n$, showing the theoretical calculation from Equation~\ref{eq_gauss_gain} as well as a Monte Carlo estimate for comparison. Note that the $x$-axis is on the $\log_{10}$ scale.}
\label{Fig_gauss_gain}
\end{center}
\end{figure}

If we assume the  $\{ X_i : i = 1, \dots, n\}$ are drawn from a standard Normal distribution with mean zero and standard deviation one, Equation~\ref{eq_infogain} reduces to
\[
G_n = \max_{i=1,\dots,n} |X_i|.
\]
We now seek to derive the expected information gain $E[G_n]$ for the Gaussian case. The cumulative distribution function of $G_n$~is
\[
\begin{array}{rclr}
   P(G_n \leq x)  & = & P(|X_1| \leq x, \dots, |X_n| \leq x) \hskip5ex & \text{(the max is $\leq x$ if and only if all values are $\leq x$)} \\[1.5ex]
     & = &\prod_{i=1}^n P(|X_i| \leq x)  & \text{(by independence)} \\[1ex]
     & = & \Big( P(|X_i| \leq x) \Big)^n & \text{(because identically distributed).}
\end{array}
\]
Also,
\[
\begin{array}{rcl}
   P(|X_i| \leq x) & =  & P(-x \leq X_i \leq x) \\[1.5ex]
     &  =  & P(X_i \leq x) - P(X_i \leq -x)  \\[1.5ex]
     &  =  & P(X_i \leq x) - (1-P(X_i \leq x)) \\[1.5ex]
     &  =  & 2P(X_i \leq x) - 1 \equiv 2\Phi(x) -1
\end{array}
\]
where $\Phi(\cdot)$ is the cumulative distribution function of the standard normal. Hence, for Gaussian data, the cumulative distribution function of $G_n$ is
\[
P(G_n \leq x) =  \Big(2\Phi(x) - 1\Big)^n.
\]
To calculate the expected gain, $E[G_n]$, we need to calculate the probability density function of $G_n$:
\[
f_n(x) = \frac{d}{dx} \Big(2\Phi(x) - 1\Big)^n = 2n \phi(x) \Big(2\Phi(x) - 1\Big)^{n-1},
\]
where $\phi(x) = \frac{d}{dx}\Phi(x)$ is the probability density function of the standard normal. Then
\begin{equation} \label{eq_gauss_gain}
E[G_n] = \int_0^\infty x f_n(x) dx,
\end{equation}
which can be solved with numerical integration.
A plot of the expected gain for Gaussian data is shown in Figure~\ref{Fig_gauss_gain} for values of $n$ ranging from $10$ to $100,000$. As a sanity check, Figure~\ref{Fig_gauss_gain} also shows a Monte Carlo estimate of the expected Gain for Gaussian data.

\subsubsection{Information gain for HENS output} 
\label{app:subsec_calcIG}

For the HENS output, we calculate the expected information gain $\widehat{G}^{\text{\tiny{HENS}}}_{n}$ in a Monte Carlo sense: for $r = 1, \dots, 2000$ and a given~$n$, 
\begin{enumerate}
    \item Randomly sample $n$ values from $\{X_i: i = 1, \dots, N\}$.
    \item Calculate $G_n(r)$ from Equation~\ref{eq_infogain}.
\end{enumerate}
Then $\widehat{G}^{\text{\tiny{HENS}}}_{n} = \frac{1}{2000}\sum_{r=1}^{2000} G_n(r)$.

\subsection{Large-sample behavior of ensemble statistics}
\label{app:LargeSample}

We derive a theory for the large-sample behavior of seven ensemble statistics:
\begin{enumerate}
    \item[1.] Mean
    \item[2.] Standard deviation
    \item[3-7.] 100$\alpha$ Percentiles for $\alpha = 0.001, 0.1, 0.5, 0.9, 0.999$ (note that $\alpha=0.5$ corresponds to the median).
\end{enumerate}
Specifically, we derive the uncertainty in each of these statistics as a function of the ensemble size $n$.

\subsubsection{Analytic uncertainty in ensemble statistics}
\label{app:AnalyticUncertainty}

To derive the theoretical or \textbf{analytic uncertainty} in various statistics, we assume that the HENS output arises from a Normal distribution with known ``population'' mean and standard deviation calculated from the full ensemble, i.e.,
\[
X_i \stackrel{\text{IID}}{\sim} N(\mu, \sigma^2),
\]
where 
\[
\mu = \overline{X} \equiv \frac{1}{N}\sum_{i=1}^N X_i, \hskip5ex \sigma^2 = \frac{1}{N-1}\sum_{i=1}^N(X_i - \overline{X})^2
\]

\vskip2ex
\noindent \textbf{Distribution of sample mean.} For a sample of size $n$, statistical theory says that the sampling distribution of $\overline{X}_n = \frac{1}{n}\sum_{i=1}^n X_i$ is
\[
\overline{X}_n \sim \mathcal{N}(\mu, \frac{\sigma^2}{n}).
\]
Hence, the analytic uncertainty in the sample mean is
\begin{equation} \label{eq_mean_unc}
SD[\overline{X}_n] = \frac{\sigma}{\sqrt{n}}.
\end{equation}

\vskip2ex
\noindent \textbf{Distribution of sample standard deviation.} For a sample of size $n$, statistical theory says that the sampling distribution of $S_n = \sqrt{ \frac{1}{n}\sum_{i=1}^n (X_i-\overline{X}_n)^2}$ is
\[
f_S(x) = 2 \frac{\left( \frac{n}{2\sigma^2}\right)^{(n-1)/2}}{\Gamma\big((n-1)/2\big)} \exp \left\{ -nx^2 / (2\sigma^2) \right\}  x^{n-2},
\]
where $\Gamma(\cdot)$ is the gamma function. The analytic uncertainty in the sample standard deviation is
\begin{equation} \label{eq_sd_unc}
SD[S_n] = \sqrt{\frac{\sigma^2}{n} \left[n - 1 - 2\frac{\Gamma^2(n/2)}{\Gamma^2(\frac{n-1}{2})} \right]}
\xrightarrow[n \to \infty]{} \frac{\sigma}{\sqrt{n}}.
\end{equation}

\vskip2ex
\noindent \textbf{Distribution of sample percentiles.} Define the \textit{order statistics} to be the sample arranged in increasing order, i.e.,
\[
X_{(1)} \leq X_{(2)} \leq \cdots \leq X_{(n)}.
\]
Statistical theory tells us that the probability density function of the $j^\text{th}$ order statistic, $X_{(j)}, j = 1,\dots,n$, is
\begin{equation} \label{eq_os_pdf}
f_{(j)}(x) = \frac{n!}{(j-1)!(n-j)!} f(x) \left[ F(x)\right]^{j-1} \left[1- F(x)\right]^{n-j}.
\end{equation}
For large enough $n$, the $100\alpha^\text{th}$ sample percentile from a sample of size $n$, denoted $X_n(\alpha)$, can be well-approximated by 
\[
X_n(\alpha) = X_{([n\alpha])},
\]
where $[y]$ is the nearest integer to $y$. We can use numerical integration to calculate the expected value of $X_n(\alpha)$ as well as its uncertainty:
\[
\text{Expected value: } \hskip3ex E[X_n(\alpha)] = \int_{-\infty}^\infty x f_{([n\alpha])}(x) dx
\]
\[
\text{Second moment: } \hskip3ex E[X_n(\alpha)^2] = \int_{-\infty}^\infty x^2 f_{([n\alpha])}(x) dx
\]
\begin{equation}\label{eq_unc}
    \text{Uncertainty: } \hskip3ex SD[X_n(\alpha)] = \sqrt{E[X_n(\alpha)^2] - E[X_n(\alpha)]^2 }.
\end{equation}

\begin{figure}[!t]
\begin{center}
\includegraphics[trim={0 0 0 0mm}, clip, width = \textwidth]{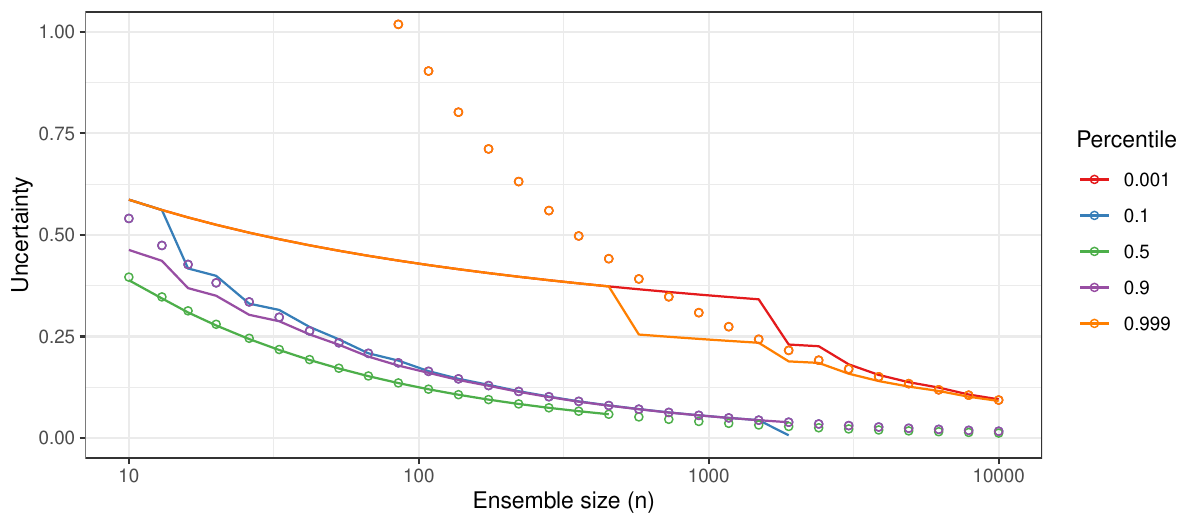}
\caption{Uncertainty in empirical percentiles from data drawn from a standard normal distribution, as calculated in Equation~\ref{eq_unc} (solid lines). Approximate uncertainties from Equation~\ref{eq_unc_apprx} are shown with empty circles. Note that the $x$-axis is on the $\log_{10}$ scale.}
\label{Fig_percentile_unc}
\end{center}
\end{figure}

\vskip2ex
\noindent Two points should be noted: first, in practice, calculation of the probability density function $f_{([n\alpha])}(x)$ can be problematic when $n$ is very large and $\alpha$ is not too close to zero and not too close to one because of the factorial terms in $\frac{n!}{(j-1)!(n-j)!}$ (i.e.,~this factorial term will be computationally equal to $\infty$). Second, if $n$ is small and $\alpha$ is close to one, there will be significant differences between $X_n(\alpha)$ (the percentile) and $X_{([n\alpha])}$ (the order statistic) due to rounding $n\alpha$. As such, the uncertainty in Equation~\ref{eq_unc} may not be representative of the actual uncertainty in the percentile. For both reasons, we turn to asymptotic theory to approximate $SD[X_n(\alpha)]$ when $n$ is very large using the central limit theorem. As the sample size $n \rightarrow \infty$, one can show that
\begin{equation} \label{eq_unc_apprx}
    SD[X_n(\alpha)] \approx 
\frac{1}{\sqrt{n}} \sqrt{\frac{\alpha(1-\alpha)}{f(F^{-1}(\alpha))^2}}.
\end{equation}
Note that we can calculate this explicitly when we know the underlying $f(\cdot)$ and $F(\cdot)$ from which the $\{X_i\}$ are drawn. For example, again consider the case where the random variables $\{X_i\}$ are drawn from a standard Normal distribution. Figure~\ref{Fig_percentile_unc} shows the uncertainty as a function of $n$ for five different percentiles. The non-smooth trajectories are due to the process of rounding $n\alpha$ when $n$ is small and $\alpha$ is close to one (as noted above). Also, the combinatorial term prevents us from calculating the expectation for large $n$ and $0.001<\alpha<0.999$.

\vskip2ex
\noindent In Figure~\ref{Fig_percentile_unc}, we see that the approximate uncertainties (circles) agree quite well with the true values for the median even for $n = 10$. Note, however, that we cannot calculate the analytic uncertainty in the median for $n>500$. For the 0.1 and 0.9 quantiles, the uncertainties align for $n \geq 100$; for the 0.001 and 0.999 quantiles, the uncertainties do not align until $n \geq 500$. Hence, for results, we will show the analytic uncertainties when possible (i.e., when they can be calculated as numerically finite) and leverage the CLT uncertainties otherwise.

\subsubsection{Extreme value theory for ``extreme'' percentiles}
\label{app:EVTPercentile}

\noindent In practice, one does not know the ``true'' distribution ($f(\cdot)$ and $F(\cdot)$) from which data are drawn. Since the density in Equation~\ref{eq_os_pdf} involves the CDF raised to the power of $n$ and $j$, any errors in estimating the CDF will compound for large $n$, making empirical estimates untenable. Furthermore, we often wish to extrapolate to percentiles beyond $1-1/n$ when $n$ is not too large, which is effectively impossible to do empirically. For both reasons, for ``extreme'' percentiles (those that are very close to zero or very close to one) we turn to extreme value theory and approximate asymptotic uncertainties.

\vskip2ex
\noindent \textbf{Estimating percentiles (i.e., return levels) using the Generalized Pareto distribution.} For a sufficiently large threshold~$u$, extreme value theory tells us that the cumulative distribution function $F(y)$ of $Y = X-u$ (conditional on $X>u$) is the Generalized Pareto Distribution (GPD):
\[
F(y) = 1 - P(X-u > y | X>u) = \left\{ \begin{array}{ll}
    1 - \left( 1 + \frac{\xi(y - u)}{\sigma} \right)^{-1/\xi} & \text{for } \xi \neq 0 \\
    1 - \exp\left\{- \frac{(y - u)}{\sigma} \right\} & \text{for } \xi = 0.
\end{array} \right.
\]
From the CDF, it follows that, for $x>u$, 
\[
P(X>x | X>u) = \left( 1 + \xi\left[\frac{x-u}{\sigma}\right] \right)^{-1/\xi};
\]
hence
\[
P(X>x) = P(X>u) \times P(X>x | X>u) = \theta_u \left( 1 + \xi\left[\frac{x-u}{\sigma}\right] \right)^{-1/\xi},
\]
where $\theta_u = P(X>u)$. Therefore, for $\alpha$ close to 1, the $100\alpha$ percentile of $f(x)$ 
is the solution of
\[
\frac{1}{1-\alpha} = \theta_u \left( 1 + \xi\left[\frac{X(\alpha)-u}{\sigma}\right] \right)^{-1/xi}.
\]
Re-arranging, we obtain a formula for the $100\alpha$ percentile
\begin{equation} \label{eq_gpd_per}
    X(\alpha) =  \left\{ \begin{array}{ll}
    u + \frac{\sigma}{\xi} \left[ (\theta_u/(1-\alpha))^\xi - 1 \right] & \text{for } \xi \neq 0 \\
    u + \sigma \log(\theta_u/(1-\alpha)) & \text{for } \xi = 0,
\end{array} \right.
\end{equation}
assuming $\alpha$ is sufficiently close to one. The percentile could alternatively be framed as the $m$-observation return level, where $m = 1/(1-\alpha)$.

\vskip2ex
\noindent \textbf{Central limit theorem for estimating percentiles with the generalized Pareto distribution.} 
The Central limit theorem says that maximum likelihood estimates $\widehat{\sigma}, \widehat{\xi}, \widehat{\theta}_u$ of the three parameters ${\sigma}, {\xi}, {\theta}_u$ in Equation~\ref{eq_gpd_per} have an asymptotically Normal distribution, wherein the uncertainties scale with $1/\sqrt{n}$.
Statistical theory allows us to derive the approximate standard error of an estimate of the $100\alpha$ percentile based on $n$ data points, denoted $\widehat{X}_n(\alpha)$. Noting that the estimate is a function of ${\sigma}, {\xi}, {\theta}_u$ as given in Equation~\ref{eq_gpd_per}, i.e., 
\begin{equation} \label{eq_rl}
\widehat{X}_n(\alpha) = g(\widehat{\sigma}, \widehat{\xi}, \widehat{\theta}_u),
\end{equation}
the delta method says that
\begin{equation}\label{eq_dm_se}
S_{\widehat{X}_n(\alpha)} = \sqrt{\nabla g^\top V \nabla g},
\end{equation}
where $V$ is the variance-covariance matrix of $(\widehat{\sigma}, \widehat{\xi}, \widehat{\theta}_u)$ (obtained from the Hessian of the GPD likelihood, evaluated at the maximum likelihood estimate) and 
\[
\nabla g = \left[\begin{array}{c}
     \frac{\partial}{\partial \sigma}  \\[0.5ex]
     \frac{\partial}{\partial \xi}  \\[0.5ex]
     \frac{\partial}{\partial \theta_u}       
\end{array} \right] = \left[\begin{array}{c}
     \xi^{-1}\big[(\theta_u/(1-\alpha))^\xi -1 \big] \\[0.5ex]
     -\sigma \xi^{-2}\big[(\theta_u/(1-\alpha))^\xi -1 \big] +  \sigma \xi^{-1}(\theta_u/(1-\alpha))^\xi \log(\theta_u/(1-\alpha))  \\[0.5ex]
     \sigma (1-\alpha)^{-\xi} \theta_u^{\xi-1}       
\end{array} \right].
\]
As a central limit theorem, this implies that uncertainties in $\widehat{X}_n(\alpha)$ will also scale with $1/\sqrt{n}$.

\subsubsection{Estimating ensemble statistics and uncertainties}
\label{app:EstStats}

To estimate the statistics of the ensemble and assess uncertainties as a function of ensemble size, we use Monte Carlo techniques similar to Section~\ref{app:subsec_calcIG}. We estimate all statistics empirically and, for the 0.001 and 0.999 quantiles, additionally using extreme value theory. For an arbitrary statistic $Z(\cdot)$, the empirical calculation proceeds as follows: for $r = 1, \dots, 2000$ and a given~$n$, 
\begin{enumerate}
    \item Randomly sample $n$ values from $\{X_i: i = 1, \dots, N\}$ \textbf{with replacement}, denoted ${\bf X}_{n}^r = (X_{1^*}, \dots, X_{n^*})$.
    \item Calculate $Z_n(r) \equiv Z({\bf X}_{n}^r)$. 
\end{enumerate}
Then, the Monte Carlo empirical estimate and empirical uncertainty are
\[
Z_n = \frac{1}{2000}\sum_r Z_n(r), \hskip5ex SD[Z_n] = \sqrt{\frac{1}{2000}\sum_r\left(Z_n(r) - \frac{1}{2000}\sum_r Z_n(r)\right)^2}.
\]
For the 0.001 and 0.999 quantiles, we obtain extreme value theory Monte Carlo estimates in a similar manner:
for \linebreak $r = 1, \dots, 2000$ and a given~$n$, 
\begin{enumerate}
    \item Randomly sample $n$ values from $\{X_i: i = 1, \dots, N\}$ \textbf{with replacement}, denoted ${\bf X}_{n}^r = (X_{1^*}, \dots, X_{n^*})$.
    \item Fit a GPD and calculate $\widehat{X}^r_n(\alpha)$ from Equation~\ref{eq_rl} and $S^r_{\widehat{X}_n(\alpha)}$ from Equation~\ref{eq_dm_se}.
\end{enumerate}
Then, the Monte Carlo GPD estimate and empirical uncertainty are
\[
\widehat{X}_n(\alpha) = \frac{1}{2000}\sum_r \widehat{X}^r_n(\alpha), \hskip5ex S_{\widehat{X}_n(\alpha)} = \frac{1}{2000}\sum_rS^r_{\widehat{X}_n(\alpha)}.
\]

\subsection{Aggregation}
\label{app:Aggregation}

Lastly, we want to aggregate over all 92 days in the HENS simulation for each of the variables. However, there is clear seasonality in the ensemble mean and standard deviation for most (if not all) of the variables. To obviate these systematic differences, we summarize all quantities in terms of standard deviations. The expected information gain quantity is already in units of standard deviations, so we can simply average the expected information gain calculated for each initialization date. Furthermore, the exchangeability ratio $R$ in Section~\ref{sec_exc} is also unitless and can hence be averaged over all initialization dates.

\vskip2ex
\noindent For the ensemble statistics, we normalize things as follows. All uncertainties are normalized by the analytic uncertainty calculated from the entire ensemble ($N=7,424$): $\sigma/\sqrt{N}$ for the mean, $SD[S_N]$ for the standard deviation, and $SD[X_N(\alpha)]$ for the quantiles (approximated using the central limit theory for $\alpha = 0.1, 0.5, 0.9$). The estimates are similarly normalized by first subtracting off the population quantity and then dividing by the analytic uncertainties for the entire ensemble. To summarize all days, we can now simply plot the average of the normalized quantities: Monte Carlo estimates plus and  minus the Monte Carlo uncertainties, and the zero line plus and minus the analytic uncertainties. For all plots, the $y$-axis has a convenient interpretation: the multiplicative uncertainty relative to the entire ensemble. In practice, one could choose the ensemble size $n$ based on how much larger they can tolerate the uncertainty relative to the ``smallest possible'' uncertainty from a huge ensemble of $N=7,424$.

\section{Uncertainties in EFI, ROC, and twCRPS metrics due to finite sample size}
\label{app:SamplingECDF}

The question of how uncertainties in the Extreme Forecast Index (EFI) and CRPS metrics scale with ensemble size can be answered by determining how the uncertainties in the empirical estimates of cumulative distribution functions (ECDFs), which govern all three metrics, scale with sample size $n$ relative to the "true" CDFs in the limit $n \to \infty$. The Dvoretzky–Kiefer–Wolfowitz (DKW) \citep{Massart_1990} inequality provides the mechanism for quantitative estimates of these uncertainties.

Let $F_n(x)$ denote the finite-sample estimate of one of the ECDFs in question (in EFI, ROC, or twCRPS metrics) and F(x) the true corresponding CDF.  For a sample of $n$ samples $(X_1, X_2, \ldots, X_N)$ with $X_i \in \mathbb{R}$, the expression for $F_n(x)$ is 
\[
F_n(x) = \frac{1}{n} \sum_{J = 1}^n \Theta\left(x - X_j\right)
\]
where $\Theta(x)$ is the Heaviside function.
The DKW inequality then states that the probability that the difference between $F_n(x)$ and F(x) exceeds a given $\epsilon$ for any x obeys the following constraint:

\[
 \hbox{Pr}\left(\sup_{x \in \mathbb{R}} \left| F_n(x) - F(x) \right| > \epsilon \right) \le 2\,e^{-2\,n\,\epsilon^2} \quad \forall \epsilon > 0 
 \label{eq:DKW}
 \]

 If we evaluate this at $\epsilon + d\epsilon$, we get
 \[
\hbox{Pr}\left(\sup_{x \in \mathbb{R}} \left| F_n(x) - F(x) \right| > \epsilon+d\epsilon \right) \le 2\,e^{-2\,n\,\left(\epsilon+d\epsilon\right)^2} \quad \forall \epsilon > 0 
\label{eq:DKWPerturbed}
\]

Differencing the second from the first equation gives
\[
\hbox{Pr}\left(\epsilon < \sup_{x \in \mathbb{R}} \left| F_n(x) - F(x) \right| \le \epsilon + d\epsilon \right) = d\epsilon\,\, \hbox{Pr}\left(\sup_{x \in \mathbb{R}} \left| F_n(x) - F(x) \right| = \epsilon  \right)\le -d\epsilon \frac{d\left[2\,e^{-2\,n\,\epsilon^2} \right]}{d\epsilon} \label{eq:DKWDiff}
\]
which, after dividing both sides by $d\epsilon$, yields the probability density of uncertainties in $F_n(x)$ as:
\begin{equation}
\hbox{Pr}\left(\sup_{x \in \mathbb{R}} \left| F_n(x) - F(x) \right| = \epsilon  \right) \le -\frac{d\left[2\,e^{-2\,n\,\epsilon^2} \right]}{d\epsilon}
\label{eq:DKWEquality}
\end{equation}
Let's denote the quantities that depend on $F_n$, i.e., the EFI, ROC, and twCRPS, by a generic function $g(F_n)$ of $F_n$ and denote the derivative of $g(F_n)$ with respect to $F_n$ by $g'(F_n)$.

Then the uncertainty in $g(F_n)$ due to finite sample size $n$ is
\[
\delta\,g = g'(F_n) \,\delta{F_n} 
\label{eq:ErrECDF}
\]
The expectation value $\langle \delta\,g \rangle$ is then
\[
\langle \delta\,g \rangle = \int g'(F_n) \,\delta{F_n} \,\hbox{Pr}\left(\delta{F_n}\right) d\,\delta{F_n}\,.
\label{eq:ExpValue}
\]
By setting $\epsilon = \delta{F_n}$, this expectation value may be written as:
\begin{equation}
\langle \delta\,g \rangle = g'(F_n) \int \epsilon\, \hbox{Pr}\left(\delta{F_n} = \epsilon\right) d\epsilon
\label{eq:ExpValueEpsilon}
\end{equation}

Following substitution of Eq.~\ref{eq:DKWEquality} into Eq.~\ref{eq:ExpValueEpsilon}, we get
\begin{equation}
\langle \delta\,g \rangle \le -g'(F_n) \int_0^1 \epsilon\,  \frac{d\left[2\,e^{-2\,n\,\epsilon^2} \right]}{d\epsilon}d\epsilon
\label{eq:ExpValueInt}
\end{equation}
since the DKW probability is an upper bound, because it governs the maximum absolute values of uncertainties, and since $0 \le \epsilon \le 1$.

Using the Leibniz product rule
\[
\frac{d\left(\epsilon\,q\right)}{d\epsilon} = \epsilon\,\frac{dq}{d\epsilon} + \cancelto{1}{\frac{d\epsilon}{d\epsilon}}q 
\label{eq:ProductRule}
\]
which implies
\[
 \int \epsilon\,\frac{dq}{d\epsilon} d\epsilon = \int \left[\frac{d\left(\epsilon\,q\right)}{d\epsilon} -  q \right]d\epsilon = \epsilon\,q - \int q d\epsilon \,,
 \label{eq:CalcImplication}
\]
we can set $q=2\,\exp(-2\,n\,\epsilon^2)$ and rewrite Eq.~\ref{eq:ExpValueInt} as
\[
\langle \delta\,g \rangle \le -g'(F_n) \left\lbrace \left.\left[\epsilon\, \left(2\,e^{-2\,n\,\epsilon^2}\right) \right]\right|_0^1 - \int_0^1 2\,e^{-2\,n\,\epsilon^2}d\epsilon\right\rbrace
\label{eq:ExpValueRewrite}
\]
This yields the uncertainty in $g(F_n)$ due to finite sample size $n$: 
\begin{eqnarray}
\langle \delta\,g \rangle &\le& -g'(F_n) \left\lbrace 2\,e^{-2\,n} - 2\,\left.\left[ \sqrt{\frac{\pi}{2}}\,\, \frac{\hbox{erf}\left(\sqrt{2\,n}\,\,\epsilon\right)}{2\sqrt{n}}\right]\right|_0^1\right\rbrace 
\label{eq:ExpValueErf}\\
&=& g'(F_n)\left\lbrace \sqrt{\frac{\pi}{2\,n}}\,\,\hbox{erf}\left(\sqrt{2\,n}\right)-2\,e^{-2\,n}\right\rbrace
\label{eq:ExpValueFinal}
\end{eqnarray}
In the limit of $n \gg 1$, the first terms in the asymptotic expansion of the error function are:
\[
\hbox{erf}(\sqrt{2\,n}) \rightarrow 1 - \frac{e^{-2\,n}}{\sqrt{2\,\pi\,n}}\;.
\label{eq:ErfAsympt}
\]
Following substitution of this expansion into Eq.~\ref{eq:ExpValueFinal},  the uncertainty simplifies to
\[
\lim_{n\gg 1}\langle \delta\,g \rangle \le g'(F_n)\left\lbrace \sqrt{\frac{\pi}{2\,n}}-\left(2+\frac{1}{2\,n}\right)\,\cancelto{\approx 0}{e^{-2\,n}}\quad\quad\right\rbrace
\label{eq:ExpValueAsympt}
\]
which in this limit is approximately
\[
\lim_{n\gg 1}\langle \delta\,g \rangle \le g'(F_n)\left\lbrace \sqrt{\frac{\pi}{2\,n}}\right\rbrace
\label{eq:ExpValueSQRT}
\]

\noappendix       




\appendixfigures  

\appendixtables   


\authorcontribution{Bold words correspond to Contributor Roles Taxonomy (CrediT) conventions. AM and WDC contributed equally to this work. AM, BB, NB, YC, PH, TK, JN, TAO, MR, DP, SS, and JW wrote \textbf{Software} and performed \textbf{Formal Data Analysis}.  WDC, KK, and MP \textbf{supervised} the research project. WDC, KK, and MP \textbf{Acquired Funding} for the project. WDC, KK, PH, SS, AM, and MP obtained computational \textbf{Resources} for the project. All authors contributed to the \textbf{Methodology} of the project.  WDC, AM, BB, YC, PH, KK, JN, TAO, MP, MR, SS, and JW contributed to the \textbf{Conceptualization} of the project.}

\competinginterests{At least one of the (co-)authors is a member of the editorial board of Geoscientific Model Development.} 

%

\begin{acknowledgements}
This research was supported by the Director, Office of Science, Office of Biological and Environmental Research of the U.S. Department of Energy under Contract No. DE-AC02-05CH11231 and by the Regional and Global Model Analysis Program area within the Earth and Environmental Systems Modeling Program. The research used resources of the National Energy Research Scientific Computing Center (NERSC), also supported by the Office of Science of the U.S. Department of Energy, under Contract No. DE-AC02-05CH11231. The computation for this paper was supported in part by the DOE Advanced Scientific Computing Research (ASCR) Leadership Computing Challenge (ALCC) 2023-2024 award `Huge Ensembles of Weather Extremes using the Fourier Forecasting Neural Network' to William Collins (LBNL). This research was also supported in part by the Environmental Resilience Institute, funded by Indiana University's Prepared for Environmental Change Grand Challenge initiative.

\end{acknowledgements}

\bibliographystyle{copernicus}
\bibliography{gmd_hens_ensemble.bib}

\begin{thebibliography}{78}
\providecommand{\natexlab}[1]{#1}
\providecommand{\url}[1]{\texttt{#1}}
\providecommand{\urlprefix}{}
\expandafter\ifx\csname urlstyle\endcsname\relax
  \providecommand{\doi}[1]{https://doi.org/\discretionary{}{}{}#1}\else
  \providecommand{\doi}{https://doi.org/\discretionary{}{}{}\begingroup \urlstyle{rm}\Url}\fi

\bibitem[{ESG()}]{ESGF}
\url{https://esgf.llnl.gov/}, accessed: 2024-07-31.

\bibitem[{MAR()}]{MARS}
\url{https://www.ecmwf.int/en/forecasts/access-forecasts/access-archive-datasets}, accessed: 2024-07-31.

\bibitem[{Per()}]{Perlmutter}
\url{https://docs.nersc.gov/systems/perlmutter/architecture/}, accessed: 2024-07-31.

\bibitem[{Allen et~al.(2023)Allen, Bhend, Martius, and Ziegel}]{Allen2023}
Allen, S., Bhend, J., Martius, O., and Ziegel, J.: Weighted Verification Tools to Evaluate Univariate and Multivariate Probabilistic Forecasts for High-Impact Weather Events, Weather and Forecasting, 38, 499–516, \doi{10.1175/waf-d-22-0161.1}, 2023.

\bibitem[{Ananthakrishnan et~al.(2014)Ananthakrishnan, Chard, Foster, and Tuecke}]{Ananthakrishnan2014}
Ananthakrishnan, R., Chard, K., Foster, I., and Tuecke, S.: Globus platform‐as‐a‐service for collaborative science applications, Concurrency and Computation: Practice and Experience, 27, 290–305, \doi{10.1002/cpe.3262}, 2014.

\bibitem[{Baño-Medina et~al.(2024)Baño-Medina, Sengupta, Watson-Parris, Hu, and Monache}]{BaoMedina2024}
Baño-Medina, J., Sengupta, A., Watson-Parris, D., Hu, W., and Monache, L.~D.: Towards calibrated ensembles of neural weather model forecasts, \doi{10.22541/essoar.171536034.43833039/v1}, 2024.

\bibitem[{Bercos‐Hickey et~al.(2022)Bercos‐Hickey, O’Brien, Wehner, Zhang, Patricola, Huang, and Risser}]{BercosHickey2022}
Bercos‐Hickey, E., O’Brien, T.~A., Wehner, M.~F., Zhang, L., Patricola, C.~M., Huang, H., and Risser, M.~D.: Anthropogenic Contributions to the 2021 Pacific Northwest Heatwave, Geophysical Research Letters, 49, \doi{10.1029/2022gl099396}, 2022.

\bibitem[{Bi et~al.(2023)Bi, Xie, Zhang, Chen, Gu, and Tian}]{Bi2023}
Bi, K., Xie, L., Zhang, H., Chen, X., Gu, X., and Tian, Q.: Accurate medium-range global weather forecasting with 3D neural networks, Nature, 619, 533–538, \doi{10.1038/s41586-023-06185-3}, 2023.

\bibitem[{Buizza and Palmer(1998)}]{Buizza1998}
Buizza, R. and Palmer, T.~N.: Impact of Ensemble Size on Ensemble Prediction, Monthly Weather Review, 126, 2503–2518, \doi{10.1175/1520-0493(1998)126<2503:ioesoe>2.0.co;2}, 1998.

\bibitem[{Caldwell et~al.(2021)Caldwell, Terai, Hillman, Keen, Bogenschutz, Lin, Beydoun, Taylor, Bertagna, Bradley, Clevenger, Donahue, Eldred, Foucar, Golaz, Guba, Jacob, Johnson, Krishna, Liu, Pressel, Salinger, Singh, Steyer, Ullrich, Wu, Yuan, Shpund, Ma, and Zender}]{Caldwell2021}
Caldwell, P.~M., Terai, C.~R., Hillman, B., Keen, N.~D., Bogenschutz, P., Lin, W., Beydoun, H., Taylor, M., Bertagna, L., Bradley, A.~M., Clevenger, T.~C., Donahue, A.~S., Eldred, C., Foucar, J., Golaz, J., Guba, O., Jacob, R., Johnson, J., Krishna, J., Liu, W., Pressel, K., Salinger, A.~G., Singh, B., Steyer, A., Ullrich, P., Wu, D., Yuan, X., Shpund, J., Ma, H., and Zender, C.~S.: Convection‐Permitting Simulations With the E3SM Global Atmosphere Model, Journal of Advances in Modeling Earth Systems, 13, \doi{10.1029/2021ms002544}, 2021.

\bibitem[{Craig et~al.(2022)Craig, Puh, Keil, Tempest, Necker, Ruiz, Weissmann, and Miyoshi}]{Craig2022}
Craig, G.~C., Puh, M., Keil, C., Tempest, K., Necker, T., Ruiz, J., Weissmann, M., and Miyoshi, T.: Distributions and convergence of forecast variables in a 1, 000‐member convection‐permitting ensemble, Quarterly Journal of the Royal Meteorological Society, 148, 2325–2343, \doi{10.1002/qj.4305}, 2022.

\bibitem[{Deser et~al.(2020)Deser, Lehner, Rodgers, Ault, Delworth, DiNezio, Fiore, Frankignoul, Fyfe, Horton, Kay, Knutti, Lovenduski, Marotzke, McKinnon, Minobe, Randerson, Screen, Simpson, and Ting}]{Deser2020}
Deser, C., Lehner, F., Rodgers, K.~B., Ault, T., Delworth, T.~L., DiNezio, P.~N., Fiore, A., Frankignoul, C., Fyfe, J.~C., Horton, D.~E., Kay, J.~E., Knutti, R., Lovenduski, N.~S., Marotzke, J., McKinnon, K.~A., Minobe, S., Randerson, J., Screen, J.~A., Simpson, I.~R., and Ting, M.: Insights from Earth system model initial-condition large ensembles and future prospects, Nature Climate Change, 10, 277–286, \doi{10.1038/s41558-020-0731-2}, 2020.

\bibitem[{Domeisen et~al.(2022)Domeisen, Eltahir, Fischer, Knutti, Perkins-Kirkpatrick, Sch\"{a}r, Seneviratne, Weisheimer, and Wernli}]{Domeisen2022}
Domeisen, D. I.~V., Eltahir, E. A.~B., Fischer, E.~M., Knutti, R., Perkins-Kirkpatrick, S.~E., Sch\"{a}r, C., Seneviratne, S.~I., Weisheimer, A., and Wernli, H.: Prediction and projection of heatwaves, Nature Reviews Earth and Environment, 4, 36–50, \doi{10.1038/s43017-022-00371-z}, 2022.

\bibitem[{Esper et~al.(2024)Esper, Torbenson, and B{\"u}ntgen}]{esper20242023}
Esper, J., Torbenson, M., and B{\"u}ntgen, U.: 2023 summer warmth unparalleled over the past 2,000 years, Nature, pp. 1--2, 2024.

\bibitem[{Eyring et~al.(2016)Eyring, Bony, Meehl, Senior, Stevens, Stouffer, and Taylor}]{Eyring2016}
Eyring, V., Bony, S., Meehl, G.~A., Senior, C.~A., Stevens, B., Stouffer, R.~J., and Taylor, K.~E.: Overview of the Coupled Model Intercomparison Project Phase 6 (CMIP6) experimental design and organization, Geoscientific Model Development, 9, 1937–1958, \doi{10.5194/gmd-9-1937-2016}, 2016.

\bibitem[{Finkel et~al.(2023)Finkel, Gerber, Abbot, and Weare}]{Finkel2023}
Finkel, J., Gerber, E.~P., Abbot, D.~S., and Weare, J.: Revealing the Statistics of Extreme Events Hidden in Short Weather Forecast Data, AGU Advances, 4, \doi{10.1029/2023av000881}, 2023.

\bibitem[{Fischer et~al.(2023)Fischer, Beyerle, Bloin-Wibe, Gessner, Humphrey, Lehner, Pendergrass, Sippel, Zeder, and Knutti}]{Fischer2023}
Fischer, E.~M., Beyerle, U., Bloin-Wibe, L., Gessner, C., Humphrey, V., Lehner, F., Pendergrass, A.~G., Sippel, S., Zeder, J., and Knutti, R.: Storylines for unprecedented heatwaves based on ensemble boosting, Nature Communications, 14, \doi{10.1038/s41467-023-40112-4}, 2023.

\bibitem[{Frame et~al.(2008)Frame, Aina, Christensen, Faull, Knight, Piani, Rosier, Yamazaki, Yamazaki, and Allen}]{Frame2008}
Frame, D., Aina, T., Christensen, C., Faull, N., Knight, S., Piani, C., Rosier, S., Yamazaki, K., Yamazaki, Y., and Allen, M.: The climate prediction .net BBC climate change experiment: design of the coupled model ensemble, Philosophical Transactions of the Royal Society A: Mathematical, Physical and Engineering Sciences, 367, 855–870, \doi{10.1098/rsta.2008.0240}, 2008.

\bibitem[{Gessner et~al.(2021)Gessner, Fischer, Beyerle, and Knutti}]{Gessner2021}
Gessner, C., Fischer, E.~M., Beyerle, U., and Knutti, R.: Very rare heat extremes: quantifying and understanding using ensemble re-initialization, Journal of Climate, p. 1–46, \doi{10.1175/jcli-d-20-0916.1}, 2021.

\bibitem[{Gneiting and Raftery(2007)}]{Gneiting2007}
Gneiting, T. and Raftery, A.~E.: Strictly Proper Scoring Rules, Prediction, and Estimation, Journal of the American Statistical Association, 102, 359–378, \doi{10.1198/016214506000001437}, 2007.

\bibitem[{Guan et~al.(2024)Guan, Arcomano, Chattopadhyay, and Maulik}]{Guan2024}
Guan, H., Arcomano, T., Chattopadhyay, A., and Maulik, R.: LUCIE: A Lightweight Uncoupled ClImate Emulator with long-term stability and physical consistency for O(1000)-member ensembles, \doi{10.48550/ARXIV.2405.16297}, 2024.

\bibitem[{Hakim and Masanam(2024)}]{Hakim2024}
Hakim, G.~J. and Masanam, S.: Dynamical Tests of a Deep-Learning Weather Prediction Model, Artificial Intelligence for the Earth Systems, \doi{10.1175/aies-d-23-0090.1}, 2024.

\bibitem[{Hazeleger et~al.(2010)Hazeleger, Severijns, Semmler, Ştefănescu, Yang, Wang, Wyser, Dutra, Baldasano, Bintanja, Bougeault, Caballero, Ekman, Christensen, van~den Hurk, Jimenez, Jones, Kållberg, Koenigk, McGrath, Miranda, van Noije, Palmer, Parodi, Schmith, Selten, Storelvmo, Sterl, Tapamo, Vancoppenolle, Viterbo, and Willén}]{Hazeleger2010}
Hazeleger, W., Severijns, C., Semmler, T., Ştefănescu, S., Yang, S., Wang, X., Wyser, K., Dutra, E., Baldasano, J.~M., Bintanja, R., Bougeault, P., Caballero, R., Ekman, A. M.~L., Christensen, J.~H., van~den Hurk, B., Jimenez, P., Jones, C., Kållberg, P., Koenigk, T., McGrath, R., Miranda, P., van Noije, T., Palmer, T., Parodi, J.~A., Schmith, T., Selten, F., Storelvmo, T., Sterl, A., Tapamo, H., Vancoppenolle, M., Viterbo, P., and Willén, U.: EC-Earth: A Seamless Earth-System Prediction Approach in Action, Bulletin of the American Meteorological Society, 91, 1357–1364, \doi{10.1175/2010bams2877.1}, 2010.

\bibitem[{Hersbach et~al.(2020)Hersbach, Bell, Berrisford, Hirahara, Horányi, Muñoz‐Sabater, Nicolas, Peubey, Radu, Schepers, Simmons, Soci, Abdalla, Abellan, Balsamo, Bechtold, Biavati, Bidlot, Bonavita, De~Chiara, Dahlgren, Dee, Diamantakis, Dragani, Flemming, Forbes, Fuentes, Geer, Haimberger, Healy, Hogan, Hólm, Janisková, Keeley, Laloyaux, Lopez, Lupu, Radnoti, de~Rosnay, Rozum, Vamborg, Villaume, and Thépaut}]{Hersbach2020}
Hersbach, H., Bell, B., Berrisford, P., Hirahara, S., Horányi, A., Muñoz‐Sabater, J., Nicolas, J., Peubey, C., Radu, R., Schepers, D., Simmons, A., Soci, C., Abdalla, S., Abellan, X., Balsamo, G., Bechtold, P., Biavati, G., Bidlot, J., Bonavita, M., De~Chiara, G., Dahlgren, P., Dee, D., Diamantakis, M., Dragani, R., Flemming, J., Forbes, R., Fuentes, M., Geer, A., Haimberger, L., Healy, S., Hogan, R.~J., Hólm, E., Janisková, M., Keeley, S., Laloyaux, P., Lopez, P., Lupu, C., Radnoti, G., de~Rosnay, P., Rozum, I., Vamborg, F., Villaume, S., and Thépaut, J.: The ERA5 global reanalysis, Quarterly Journal of the Royal Meteorological Society, 146, 1999–2049, \doi{10.1002/qj.3803}, 2020.

\bibitem[{Hu et~al.(2023)Hu, Chen, Wang, and Li}]{Hu2023}
Hu, Y., Chen, L., Wang, Z., and Li, H.: SwinVRNN: A Data‐Driven Ensemble Forecasting Model via Learned Distribution Perturbation, Journal of Advances in Modeling Earth Systems, 15, \doi{10.1029/2022ms003211}, 2023.

\bibitem[{Jeffrey et~al.(2013)Jeffrey, Rotstayn, Collier, Dravitzki, Hamalainen, Moeseneder, Wong, and Syktus}]{jeffrey2013australia}
Jeffrey, S., Rotstayn, L., Collier, M., Dravitzki, S., Hamalainen, C., Moeseneder, C., Wong, K., and Syktus, J.: Australia’s CMIP5 submission usingthe CSIRO-Mk3. 6 model, Australian Meteorological and Oceanographic Journal, 63, 1--13, 2013.

\bibitem[{Kay et~al.(2015)Kay, Deser, Phillips, Mai, Hannay, Strand, Arblaster, Bates, Danabasoglu, Edwards, Holland, Kushner, Lamarque, Lawrence, Lindsay, Middleton, Munoz, Neale, Oleson, Polvani, and Vertenstein}]{Kay2015}
Kay, J.~E., Deser, C., Phillips, A., Mai, A., Hannay, C., Strand, G., Arblaster, J.~M., Bates, S.~C., Danabasoglu, G., Edwards, J., Holland, M., Kushner, P., Lamarque, J.-F., Lawrence, D., Lindsay, K., Middleton, A., Munoz, E., Neale, R., Oleson, K., Polvani, L., and Vertenstein, M.: The Community Earth System Model (CESM) Large Ensemble Project: A Community Resource for Studying Climate Change in the Presence of Internal Climate Variability, Bulletin of the American Meteorological Society, 96, 1333–1349, \doi{10.1175/bams-d-13-00255.1}, 2015.

\bibitem[{Kelder et~al.(2022{\natexlab{a}})Kelder, Marjoribanks, Slater, Prudhomme, Wilby, Wagemann, and Dunstone}]{Kelder2022IOP}
Kelder, T., Marjoribanks, T.~I., Slater, L.~J., Prudhomme, C., Wilby, R.~L., Wagemann, J., and Dunstone, N.: An open workflow to gain insights about low‐likelihood high‐impact weather events from initialized predictions, Meteorological Applications, 29, \doi{10.1002/met.2065}, 2022{\natexlab{a}}.

\bibitem[{Kelder et~al.(2022{\natexlab{b}})Kelder, Wanders, van~der Wiel, Marjoribanks, Slater, Wilby, and Prudhomme}]{Kelder2022}
Kelder, T., Wanders, N., van~der Wiel, K., Marjoribanks, T.~I., Slater, L.~J., Wilby, R.~l., and Prudhomme, C.: Interpreting extreme climate impacts from large ensemble simulations—are they unseen or unrealistic?, Environmental Research Letters, 17, 044\,052, \doi{10.1088/1748-9326/ac5cf4}, 2022{\natexlab{b}}.

\bibitem[{Kirchmeier-Young and Zhang(2020)}]{KirchmeierYoung2020}
Kirchmeier-Young, M.~C. and Zhang, X.: Human influence has intensified extreme precipitation in North America, Proceedings of the National Academy of Sciences, 117, 13\,308–13\,313, \doi{10.1073/pnas.1921628117}, 2020.

\bibitem[{Kirchmeier-Young et~al.(2017)Kirchmeier-Young, Zwiers, and Gillett}]{KirchmeierYoung2017}
Kirchmeier-Young, M.~C., Zwiers, F.~W., and Gillett, N.~P.: Attribution of Extreme Events in Arctic Sea Ice Extent, Journal of Climate, 30, 553–571, \doi{10.1175/jcli-d-16-0412.1}, 2017.

\bibitem[{Kochkov et~al.(2023)Kochkov, Yuval, Langmore, Norgaard, Smith, Mooers, Lottes, Rasp, D\"{u}ben, Kl\"{o}wer, Hatfield, Battaglia, Sanchez-Gonzalez, Willson, Brenner, and Hoyer}]{Kochkov2023}
Kochkov, D., Yuval, J., Langmore, I., Norgaard, P., Smith, J., Mooers, G., Lottes, J., Rasp, S., D\"{u}ben, P., Kl\"{o}wer, M., Hatfield, S., Battaglia, P., Sanchez-Gonzalez, A., Willson, M., Brenner, M.~P., and Hoyer, S.: Neural General Circulation Models, \doi{10.48550/ARXIV.2311.07222}, 2023.

\bibitem[{Leach et~al.(2021)Leach, Weisheimer, Allen, and Palmer}]{Leach2021}
Leach, N.~J., Weisheimer, A., Allen, M.~R., and Palmer, T.: Forecast-based attribution of a winter heatwave within the limit of predictability, Proceedings of the National Academy of Sciences, 118, \doi{10.1073/pnas.2112087118}, 2021.

\bibitem[{Leach et~al.(2022)Leach, Watson, Sparrow, Wallom, and Sexton}]{Leach2022}
Leach, N.~J., Watson, P.~A., Sparrow, S.~N., Wallom, D.~C., and Sexton, D.~M.: Generating samples of extreme winters to support climate adaptation, Weather and Climate Extremes, 36, 100\,419, \doi{10.1016/j.wace.2022.100419}, 2022.

\bibitem[{Leach et~al.(2024)Leach, Roberts, Aengenheyster, Heathcote, Mitchell, Thompson, Palmer, Weisheimer, and Allen}]{Leach2024}
Leach, N.~J., Roberts, C.~D., Aengenheyster, M., Heathcote, D., Mitchell, D.~M., Thompson, V., Palmer, T., Weisheimer, A., and Allen, M.~R.: Heatwave attribution based on reliable operational weather forecasts, Nature Communications, 15, \doi{10.1038/s41467-024-48280-7}, 2024.

\bibitem[{Lerch et~al.(2017)Lerch, Thorarinsdottir, Ravazzolo, and Gneiting}]{Lerch2017}
Lerch, S., Thorarinsdottir, T.~L., Ravazzolo, F., and Gneiting, T.: Forecaster’s Dilemma: Extreme Events and Forecast Evaluation, Statistical Science, 32, \doi{10.1214/16-sts588}, 2017.

\bibitem[{Leutbecher(2018)}]{Leutbecher2018}
Leutbecher, M.: Ensemble size: How suboptimal is less than infinity?, Quarterly Journal of the Royal Meteorological Society, 145, 107–128, \doi{10.1002/qj.3387}, 2018.

\bibitem[{Leutbecher and Palmer(2008)}]{Leutbecher2008}
Leutbecher, M. and Palmer, T.: Ensemble forecasting, Journal of Computational Physics, 227, 3515–3539, \doi{10.1016/j.jcp.2007.02.014}, 2008.

\bibitem[{Li et~al.(2024)Li, Carver, Lopez-Gomez, Sha, and Anderson}]{Li2024}
Li, L., Carver, R., Lopez-Gomez, I., Sha, F., and Anderson, J.: Generative emulation of weather forecast ensembles with diffusion models, Science Advances, 10, \doi{10.1126/sciadv.adk4489}, 2024.

\bibitem[{Longmate et~al.(2023)Longmate, Risser, and Feldman}]{Longmate_2023}
Longmate, J.~M., Risser, M.~D., and Feldman, D.~R.: {Prioritizing the selection of CMIP6 model ensemble members for downscaling projections of CONUS temperature and precipitation}, Climate Dynamics, 61, 5171–5197, \doi{10.1007/s00382-023-06846-z}, 2023.

\bibitem[{Lu and Romps(2022)}]{Lu2022}
Lu, Y.-C. and Romps, D.~M.: Extending the Heat Index, Journal of Applied Meteorology and Climatology, 61, 1367–1383, \doi{10.1175/jamc-d-22-0021.1}, 2022.

\bibitem[{Maher et~al.(2019)Maher, Milinski, Suarez‐Gutierrez, Botzet, Dobrynin, Kornblueh, Kr\"{o}ger, Takano, Ghosh, Hedemann, Li, Li, Manzini, Notz, Putrasahan, Boysen, Claussen, Ilyina, Olonscheck, Raddatz, Stevens, and Marotzke}]{Maher2019}
Maher, N., Milinski, S., Suarez‐Gutierrez, L., Botzet, M., Dobrynin, M., Kornblueh, L., Kr\"{o}ger, J., Takano, Y., Ghosh, R., Hedemann, C., Li, C., Li, H., Manzini, E., Notz, D., Putrasahan, D., Boysen, L., Claussen, M., Ilyina, T., Olonscheck, D., Raddatz, T., Stevens, B., and Marotzke, J.: The Max Planck Institute Grand Ensemble: Enabling the Exploration of Climate System Variability, Journal of Advances in Modeling Earth Systems, 11, 2050–2069, \doi{10.1029/2019ms001639}, 2019.

\bibitem[{Mahesh et~al.(2023)Mahesh, O’Brien, Loring, Elbashandy, Boos, and Collins}]{Mahesh2023}
Mahesh, A., O’Brien, T., Loring, B., Elbashandy, A., Boos, W., and Collins, W.: Identifying Atmospheric Rivers and their Poleward Latent Heat Transport with Generalizable Neural Networks: ARCNNv1, \doi{10.5194/egusphere-2023-763}, 2023.

\bibitem[{Mahesh et~al.(2024)Mahesh, Collins, Bonev, Brenowitz, Cohen, Elms, Harrington, Kashinath, Kurth, North, OBrien, Pritchard, Pruitt, Risser, Subramanian, and Willard}]{MaheshPartI2024}
Mahesh, A., Collins, W., Bonev, B., Brenowitz, N., Cohen, Y., Elms, J., Harrington, P., Kashinath, K., Kurth, T., North, J., OBrien, T., Pritchard, M., Pruitt, D., Risser, M., Subramanian, S., and Willard, J.: Huge Ensembles Part I: Design of Ensemble Weather Forecasts using Spherical Fourier Neural Operators, \doi{10.48550/ARXIV.2408.03100}, 2024.

\bibitem[{Mamalakis et~al.(2022)Mamalakis, Ebert-Uphoff, and Barnes}]{Mamalakis2022}
Mamalakis, A., Ebert-Uphoff, I., and Barnes, E.~A.: Neural network attribution methods for problems in geoscience: A novel synthetic benchmark dataset, Environmental Data Science, 1, \doi{10.1017/eds.2022.7}, 2022.

\bibitem[{Mankin et~al.(2020)Mankin, Lehner, Coats, and McKinnon}]{Mankin2020}
Mankin, J.~S., Lehner, F., Coats, S., and McKinnon, K.~A.: The Value of Initial Condition Large Ensembles to Robust Adaptation Decision‐Making, Earth’s Future, 8, \doi{10.1029/2020ef001610}, 2020.

\bibitem[{Massart(1990)}]{Massart_1990}
Massart, P.: {The Tight Constant in the Dvoretzky-Kiefer-Wolfowitz Inequality}, The Annals of Probability, 18, \doi{10.1214/aop/1176990746}, 1990.

\bibitem[{McCulloch and Neuhaus(2011)}]{McCulloch_2011}
McCulloch, C.~E. and Neuhaus, J.~M.: {Misspecifying the Shape of a Random Effects Distribution: Why Getting It Wrong May Not Matter}, Statistical Science, 26, \doi{10.1214/11-sts361}, 2011.

\bibitem[{McKinnon et~al.(2017)McKinnon, Poppick, Dunn-Sigouin, and Deser}]{McKinnon2017}
McKinnon, K.~A., Poppick, A., Dunn-Sigouin, E., and Deser, C.: An “Observational Large Ensemble” to Compare Observed and Modeled Temperature Trend Uncertainty due to Internal Variability, Journal of Climate, 30, 7585–7598, \doi{10.1175/jcli-d-16-0905.1}, 2017.

\bibitem[{Milinski et~al.(2020)Milinski, Maher, and Olonscheck}]{Milinski2020}
Milinski, S., Maher, N., and Olonscheck, D.: How large does a large ensemble need to be?, Earth System Dynamics, 11, 885–901, \doi{10.5194/esd-11-885-2020}, 2020.

\bibitem[{Millin and Furtado(2022)}]{Millin2022}
Millin, O.~T. and Furtado, J.~C.: The Role of Wave Breaking in the Development and Subseasonal Forecasts of the February 2021 Great Plains Cold Air Outbreak, Geophysical Research Letters, 49, \doi{10.1029/2022gl100835}, 2022.

\bibitem[{Miranda et~al.(2023)Miranda, Lizana, Sparrow, Zachau-Walker, Watson, Wallom, Khosla, and McCulloch}]{Miranda2023}
Miranda, N.~D., Lizana, J., Sparrow, S.~N., Zachau-Walker, M., Watson, P. A.~G., Wallom, D. C.~H., Khosla, R., and McCulloch, M.: Change in cooling degree days with global mean temperature rise increasing from 1.5 °C to 2.0 °C, Nature Sustainability, 6, 1326–1330, \doi{10.1038/s41893-023-01155-z}, 2023.

\bibitem[{Mo et~al.(2022)Mo, Lin, and Vitart}]{Mo2022}
Mo, R., Lin, H., and Vitart, F.: An anomalous warm-season trans-Pacific atmospheric river linked to the 2021 western North America heatwave, Communications Earth and Environment, 3, \doi{10.1038/s43247-022-00459-w}, 2022.

\bibitem[{Palmer(2002)}]{Palmer2002}
Palmer, T.~N.: The economic value of ensemble forecasts as a tool for risk assessment: From days to decades, Quarterly Journal of the Royal Meteorological Society, 128, 747–774, \doi{10.1256/0035900021643593}, 2002.

\bibitem[{Pathak et~al.(2022)Pathak, Subramanian, Harrington, Raja, Chattopadhyay, Mardani, Kurth, Hall, Li, Azizzadenesheli, Hassanzadeh, Kashinath, and Anandkumar}]{Pathak2022}
Pathak, J., Subramanian, S., Harrington, P., Raja, S., Chattopadhyay, A., Mardani, M., Kurth, T., Hall, D., Li, Z., Azizzadenesheli, K., Hassanzadeh, P., Kashinath, K., and Anandkumar, A.: FourCastNet: A Global Data-driven High-resolution Weather Model using Adaptive Fourier Neural Operators, \doi{10.48550/ARXIV.2202.11214}, 2022.

\bibitem[{Philip et~al.(2022)Philip, Kew, van Oldenborgh, Anslow, Seneviratne, Vautard, Coumou, Ebi, Arrighi, Singh, van Aalst, Pereira~Marghidan, Wehner, Yang, Li, Schumacher, Hauser, Bonnet, Luu, Lehner, Gillett, Tradowsky, Vecchi, Rodell, Stull, Howard, and Otto}]{Philip2022}
Philip, S.~Y., Kew, S.~F., van Oldenborgh, G.~J., Anslow, F.~S., Seneviratne, S.~I., Vautard, R., Coumou, D., Ebi, K.~L., Arrighi, J., Singh, R., van Aalst, M., Pereira~Marghidan, C., Wehner, M., Yang, W., Li, S., Schumacher, D.~L., Hauser, M., Bonnet, R., Luu, L.~N., Lehner, F., Gillett, N., Tradowsky, J.~S., Vecchi, G.~A., Rodell, C., Stull, R.~B., Howard, R., and Otto, F. E.~L.: Rapid attribution analysis of the extraordinary heat wave on the Pacific coast of the US and Canada in June 2021, Earth System Dynamics, 13, 1689–1713, \doi{10.5194/esd-13-1689-2022}, 2022.

\bibitem[{Price et~al.(2023)Price, Sanchez-Gonzalez, Alet, Ewalds, El-Kadi, Stott, Mohamed, Battaglia, Lam, and Willson}]{Price2023}
Price, I., Sanchez-Gonzalez, A., Alet, F., Ewalds, T., El-Kadi, A., Stott, J., Mohamed, S., Battaglia, P., Lam, R., and Willson, M.: GenCast: Diffusion-based ensemble forecasting for medium-range weather, \doi{10.48550/ARXIV.2312.15796}, 2023.

\bibitem[{Richardson(2001)}]{Richardson2001}
Richardson, D.~S.: Measures of skill and value of ensemble prediction systems, their interrelationship and the effect of ensemble size, Quarterly Journal of the Royal Meteorological Society, 127, 2473–2489, \doi{10.1002/qj.49712757715}, 2001.

\bibitem[{Rodgers et~al.(2015)Rodgers, Lin, and Fr\"{o}licher}]{Rodgers2015}
Rodgers, K.~B., Lin, J., and Fr\"{o}licher, T.~L.: Emergence of multiple ocean ecosystem drivers in a large ensemble suite with an Earth system model, Biogeosciences, 12, 3301–3320, \doi{10.5194/bg-12-3301-2015}, 2015.

\bibitem[{Runge et~al.(2019)Runge, Bathiany, Bollt, Camps-Valls, Coumou, Deyle, Glymour, Kretschmer, Mahecha, Muñoz-Marí, van Nes, Peters, Quax, Reichstein, Scheffer, Sch\"{o}lkopf, Spirtes, Sugihara, Sun, Zhang, and Zscheischler}]{Runge2019}
Runge, J., Bathiany, S., Bollt, E., Camps-Valls, G., Coumou, D., Deyle, E., Glymour, C., Kretschmer, M., Mahecha, M.~D., Muñoz-Marí, J., van Nes, E.~H., Peters, J., Quax, R., Reichstein, M., Scheffer, M., Sch\"{o}lkopf, B., Spirtes, P., Sugihara, G., Sun, J., Zhang, K., and Zscheischler, J.: Inferring causation from time series in Earth system sciences, Nature Communications, 10, \doi{10.1038/s41467-019-10105-3}, 2019.

\bibitem[{Sanderson et~al.(2015)Sanderson, Oleson, Strand, Lehner, and O’Neill}]{Sanderson2015}
Sanderson, B.~M., Oleson, K.~W., Strand, W.~G., Lehner, F., and O’Neill, B.~C.: A new ensemble of GCM simulations to assess avoided impacts in a climate mitigation scenario, Climatic Change, 146, 303–318, \doi{10.1007/s10584-015-1567-z}, 2015.

\bibitem[{Scher and Messori(2021)}]{Scher2021}
Scher, S. and Messori, G.: Ensemble Methods for Neural Network‐Based Weather Forecasts, Journal of Advances in Modeling Earth Systems, 13, \doi{10.1029/2020ms002331}, 2021.

\bibitem[{Schneider et~al.(2024)Schneider, Leung, and Wills}]{Schneider2024}
Schneider, T., Leung, L.~R., and Wills, R. C.~J.: Opinion: Optimizing climate models with process knowledge, resolution, and artificial intelligence, Atmospheric Chemistry and Physics, 24, 7041–7062, \doi{10.5194/acp-24-7041-2024}, 2024.

\bibitem[{Siegert et~al.(2019)Siegert, Ferro, Stephenson, and Leutbecher}]{Siegert2019}
Siegert, S., Ferro, C. A.~T., Stephenson, D.~B., and Leutbecher, M.: The ensemble‐adjusted Ignorance Score for forecasts issued as normal distributions, Quarterly Journal of the Royal Meteorological Society, 145, 129–139, \doi{10.1002/qj.3447}, 2019.

\bibitem[{Steadman(1979)}]{Steadman1979}
Steadman, R.~G.: The Assessment of Sultriness. Part I: A Temperature-Humidity Index Based on Human Physiology and Clothing Science, Journal of Applied Meteorology, 18, 861–873, \doi{10.1175/1520-0450(1979)018<0861:taospi>2.0.co;2}, 1979.

\bibitem[{Sun et~al.(2018)Sun, Alexander, and Deser}]{Sun2018}
Sun, L., Alexander, M., and Deser, C.: Evolution of the Global Coupled Climate Response to Arctic Sea Ice Loss during 1990–2090 and Its Contribution to Climate Change, Journal of Climate, 31, 7823–7843, \doi{10.1175/jcli-d-18-0134.1}, 2018.

\bibitem[{Swain et~al.(2020)Swain, Wing, Bates, Done, Johnson, and Cameron}]{Swain2020}
Swain, D.~L., Wing, O. E.~J., Bates, P.~D., Done, J.~M., Johnson, K.~A., and Cameron, D.~R.: Increased Flood Exposure Due to Climate Change and Population Growth in the United States, Earth’s Future, 8, \doi{10.1029/2020ef001778}, 2020.

\bibitem[{Thompson et~al.(2017)Thompson, Dunstone, Scaife, Smith, Slingo, Brown, and Belcher}]{Thompson2017}
Thompson, V., Dunstone, N.~J., Scaife, A.~A., Smith, D.~M., Slingo, J.~M., Brown, S., and Belcher, S.~E.: {High risk of unprecedented {UK} rainfall in the current climate}, Nature Communications, 8, \doi{10.1038/s41467-017-00275-3}, 2017.

\bibitem[{Vonich and Hakim(2024)}]{Vonich2024}
Vonich, P.~T. and Hakim, G.~J.: Predictability Limit of the 2021 Pacific Northwest Heatwave from Deep-Learning Sensitivity Analysis, \doi{10.48550/ARXIV.2406.05019}, 2024.

\bibitem[{Webber et~al.(2019)Webber, Plotkin, O’Neill, Abbot, and Weare}]{Webber2019}
Webber, R.~J., Plotkin, D.~A., O’Neill, M.~E., Abbot, D.~S., and Weare, J.: Practical rare event sampling for extreme mesoscale weather, Chaos: An Interdisciplinary Journal of Nonlinear Science, 29, \doi{10.1063/1.5081461}, 2019.

\bibitem[{Weyn et~al.(2019)Weyn, Durran, and Caruana}]{Weyn2019}
Weyn, J.~A., Durran, D.~R., and Caruana, R.: Can Machines Learn to Predict Weather? Using Deep Learning to Predict Gridded 500‐hPa Geopotential Height From Historical Weather Data, Journal of Advances in Modeling Earth Systems, 11, 2680–2693, \doi{10.1029/2019ms001705}, 2019.

\bibitem[{Weyn et~al.(2021)Weyn, Durran, Caruana, and Cresswell‐Clay}]{Weyn2021}
Weyn, J.~A., Durran, D.~R., Caruana, R., and Cresswell‐Clay, N.: Sub‐Seasonal Forecasting With a Large Ensemble of Deep‐Learning Weather Prediction Models, Journal of Advances in Modeling Earth Systems, 13, \doi{10.1029/2021ms002502}, 2021.

\bibitem[{Wilks and Hamill(1995)}]{Wilks1995}
Wilks, D.~S. and Hamill, T.~M.: Potential Economic Value of Ensemble-Based Surface Weather Forecasts, Monthly Weather Review, 123, 3565–3575, \doi{10.1175/1520-0493(1995)123<3565:pevoeb>2.0.co;2}, 1995.

\bibitem[{Ye et~al.(2024)Ye, Woollings, Sparrow, Watson, and Screen}]{Ye2024}
Ye, K., Woollings, T., Sparrow, S.~N., Watson, P. A.~G., and Screen, J.~A.: Response of winter climate and extreme weather to projected Arctic sea-ice loss in very large-ensemble climate model simulations, npj Climate and Atmospheric Science, 7, \doi{10.1038/s41612-023-00562-5}, 2024.

\bibitem[{Zamo and Naveau(2017)}]{Zamo2017}
Zamo, M. and Naveau, P.: Estimation of the Continuous Ranked Probability Score with Limited Information and Applications to Ensemble Weather Forecasts, Mathematical Geosciences, 50, 209–234, \doi{10.1007/s11004-017-9709-7}, 2017.

\bibitem[{Zhang et~al.(2024)Zhang, Risser, Wehner, and O’Brien}]{Zhang2024}
Zhang, L., Risser, M.~D., Wehner, M.~F., and O’Brien, T.~A.: Leveraging Extremal Dependence to Better Characterize the 2021 Pacific Northwest Heatwave, Journal of Agricultural, Biological and Environmental Statistics, \doi{10.1007/s13253-024-00636-8}, 2024.

\bibitem[{Zhang and Boos(2023)}]{Zhang2023}
Zhang, Y. and Boos, W.~R.: An upper bound for extreme temperatures over midlatitude land, Proceedings of the National Academy of Sciences, 120, \doi{10.1073/pnas.2215278120}, 2023.

\bibitem[{Zhong et~al.(2024)Zhong, Chen, Li, Feng, and Lu}]{Zhong2024}
Zhong, X., Chen, L., Li, H., Feng, J., and Lu, B.: FuXi-ENS: A machine learning model for medium-range ensemble weather forecasting, \doi{10.48550/ARXIV.2405.05925}, 2024.

\end{thebibliography}

\end{document}